\documentclass[12pt]{article}

\RequirePackage[OT1]{fontenc}
\RequirePackage{amsthm,amsmath,amssymb,amstext,graphicx,psfrag}

\usepackage{hero}
\usepackage{turnstile}
\usepackage{psfrag,epsfig,multimedia}
\usepackage{graphicx}
\usepackage{amssymb,amsmath,amsfonts}
\usepackage{enumerate}

\usepackage{setspace}

\usepackage[tmargin=1in, bmargin=1in, rmargin=1in, lmargin=1in]{geometry}
\usepackage{setspace}
\usepackage{chapterbib}
\usepackage{fancyhdr}

\numberwithin{equation}{section}
\theoremstyle{plain}

\renewcommand{\r}{\mbox{\rm r}}
\def\ra{\mbox{\rm a}}

\title{
Large Scale Correlation Screening 
}

\author{
Alfred Hero\\
Department of EECS, BME and Statistics\\
University of Michigan, Ann Arbor, MI 48109
\and
Bala Rajaratnam\\
Department of Statistics\\
Stanford University, Stanford, CA 94305
}

\begin{document}

%
%
%
%

\maketitle

%
%

\noindent {\bf Note that this arxiv version is an updated version of the technical report uploaded on Arxiv on February 06, 2011. An earlier version of this report submitted for publication on March 31, 2010 is also available on request. }

\begin{abstract}
This paper treats the problem of screening for variables with high correlations
in high dimensional data in which there can be many fewer samples than variables.
We focus on threshold-based correlation screening methods for three related applications:
screening for variables with large correlations within a single treatment (autocorrelation
screening); screening for variables with large cross-correlations over two
treatments (cross-correlation screening); screening for variables that have persistently
large auto-correlations over two treatments (persistent-correlation screening). The novelty
of correlation screening is that it identifies a smaller number of \mbox{{\it variables}} which
are highly correlated with others, as compared to identifying a number of correlation
\mbox{{\it parameters}}. Correlation screening suffers from a phase transition phenomenon: as
the correlation threshold decreases the number of discoveries increases abruptly. We
obtain asymptotic expressions for the mean number of discoveries and the phase transition
thresholds as a function of the number of samples, the number of variables, and
the joint sample distribution. We also show that under a weak dependency condition
the number of discoveries is dominated by a Poisson random variable giving an
asymptotic expression for the false positive rate. The correlation screening approach
bears  tremendous dividends in terms of the type and strength of the asymptotic results
that can be obtained. It also overcomes some of the major hurdles faced by existing
methods in the literature as correlation screening is naturally scalable to high dimension. Numerical
results  strongly validate the theory that is presented in this paper.
We illustrate the application of the correlation screening methodology on a large scale
gene-expression dataset, revealing a few influential variables that
exhibit a significant amount of correlation over multiple treatments.

\end{abstract}


%
%

%
%
%
%
%
%

\doublespacing
\section{Introduction}

Consider the problem of screening for variables that have significant
correlations in a large data set. Examples of such data sets are gene
expression arrays, multimedia databases, multivariate financial time
series, and traffic over the Internet. Correlation screening can be
used to discover a small number of variables that are highly
correlated or whose correlations have distinct patterns, or motifs,
that are not likely to occur by chance. Indeed, filtering out all but
the highest sample correlations may be the only practical way to
examine dependencies in massive datasets where computational
limitations prevent the experimenter from evaluating all sample
correlations. As an example, in multi-chip gene expression data the
number of pairwise correlations can be in the billions.

Thresholding the sample correlation matrix is an attractive screening
method due to its simplicity.  However, the threshold must be chosen
with care due to the existence of an abrupt phase transition
phenomenon controlling the number of discoveries.  When the
correlation threshold falls below a critical point the number of
discoveries abruptly and rapidly increases, even when the variables
are uncorrelated. This critical point can be close to one when the
number $p$ of variables greatly exceeds the number $n$ of samples.
Therefore a poorly selected correlation threshold may result in an
overwhelmingly large number of discoveries.  This paper provides
theory that predicts the location of this critical point as a function
of $n$, $p$, and the joint distribution of the variables. When the
population covariance matrix is of large dimension and sparse the
theory specifies universal thresholds that do not depend on the
unknown multivariate sample density.

We distinguish between three types of screening which arise in
practical applications involving a single treatment or a pair of
treatments. Each type of screening seeks to discover variables with
the property that they are highly correlated with at least one other
variable. The first application involves screening for variables that
are highly correlated with other variables in undergoing the same
treatment. The second application is screening for variables in one
treatment that are highly correlated with variables undergoing a
different treatment. The third application is screening for variables
with high within-treatment correlation that persists over a pair of
treatments. Precise definitions are given
in Section \ref{sec:correlationscreening}. We respectively call these
three applications auto-correlation screening, cross-correlation
screening, and persistent-correlation screening. In each of these
problems the location of the phase transition critical point is
different.

For each of these three applications we index the correlation
threshold $\rho_p$ by the number of variables $p$.  We give asymptotic
conditions on the sequence $\{\rho_p\}_p$ that guarantee a finite and
non-zero mean number of discoveries. These conditions, which depend on
the number $n$ of samples, can be used to guide the selection of an
appropriate correlation threshold in practical applications. Under
these conditions we derive asymptotic expressions for the mean number
of discoveries.  These expressions depend on a Bhattacharyya
measure \cite{Basseville:SP89} of average pairwise dependency of the $p$ multivariate
U-scores defined on the $(n-2)$-dimensional hypersphere.
It is through this pairwise dependency measure that the population
covariance matrix influences the mean number of discoveries.

We establish simple achievable bounds that give insight into factors
that determine the mean number of discoveries.  These bounds involve
R\'enyi entropy \cite{Renyi:61} and other information theoretic
quantities. For example, we show that the mean discovery rate
is proportional to the order 2 R\'{e}nyi entropy of the average
marginal density of associated U-scores if and only if these scores
are independent identically distributed. Under this i.i.d. condition
the mean number of auto-correlation screening discoveries is minimized
for the case of uniformly distributed U-scores. This establishes a
minimal property of the $p$-variate spherical distribution over the
elliptical diagonal dispersion family.

Using the expressions for the mean number of discoveries we specify
the critical point $\rho_c$ of the phase transition.  As either $p$
increases or $n$ decreases $\rho_c$ approaches one, making reliable
screening impossible, and $\rho_c$ approaches this limit with rate
roughly equal to $p^{-1/n}$. In particular, for auto-correlation
screening, when $n>4$ and $p$ is large:
$\rho_c=\sqrt{1-c_n(p-1)^{-2/(n-4)}}$, where $c_n$ depends on the
aforementioned Bhattacharyya measure of average pairwise dependency of
the U-scores
and only depends weakly on $n$.

We also establish that under a weak dependency assumption the number
of discoveries is asymptotically dominated by a related Poisson random
variable. In the case of auto-correlation and cross-correlation
screening this Poisson variable is the number of positive vertex
degrees in the associated sample correlation graph.  In the case of
persistent-correlation screening the dominating Poisson variable is
the correlation of the vertex degrees in the sample correlation graphs
associated with each treatment.  The
weak dependency condition on the average U-score pairwise
distributions is satisfied for variables whose covariance matrix is
sparse or whose correlations are small.

These dominance results specify an asymptotic expression for the false
positive rate of discoveries that can be used to select the screening
threshold to control the familywise discovery rate. Familywise discovery rate has been widely used in variable selection problems. The rate function in our derived Poisson limit specifies the marginal false discovery rate associated with a particular correlation threshold.  While we do not explore it in this paper, when suitably corrected for dependency, the associated p-values might also be used to control the conditional false discovery rate. For a given pair of variables and a given screening threshold, the bias-corrected
normal approximation to the Fisher Z transformed sample correlations allows
us to approximate the minimum detectable correlation between the
variables. We give a numerical example that provides experimental
validation and illustrates the practical utility of our theoretical
predictions for large but finite $p$ and small $n$. We then apply our
method to correlation screening of a large scale Affymetrix gene
micro-array dataset for analysis of a four treatment beverage intake
experiment \cite{baty2006analysis}.

The correlation screening problem treated here is not related to inverse
covariance and covariance selection problems studied by many authors (see
\cite{Bickel&Levina:AnnStat08_1,Ledoit&Wolf:JMA04,Rothman&etal:08,Friedman&etal:07,Dey&Srinivasan:AnnStat85,Rajaratnam&etal:08} to name just a few from an increasing literature).
Unlike these authors who are interested in correlation or
covariance matrix estimation with respect to a matrix error norm, here
we are concerned with detection of a few variables with large
correlation coefficients. 
Unlike previous work in covariance selection we provide precise phase transition
thresholds that are applicable to large scale screening for correlation and persistence in single and multiple treatments. This paper is related to tests of significance for covariance and correlation matrices \cite{Hills:Biometrika69, Cameron&Eagleson:AJS85}, but our focus is correlation screening instead of testing for diagonal covariance or for other structure. Tests of diagonal covariance structure are often based on the maximum sample correlation coefficient, which has recently been studied in the large $p$ regime \cite{Jiang:04, Lietal:PTRF09,  Li&Rosalsky:06, Liuetal:08, Zhou:07}. Unlike the correlation screening results shown in this paper, these studies often impose  more stringent (Gaussian) assumptions on the joint distribution of the variables and do not consider the case of persistent maximal correlation. On the other hand, our results could be of practical value in both covariance selection and correlation tests of significance, especially when $p$ is large.

%

Correlation screening  is an effective method for discovering a few highly correlated variables when there are no response variables in the data, i.e., it is an unsupervised method. While our formulation of correlation screening does not specifically target the supervised problem of variable selection for regression, the correlation screening framework  can be applied to this setting. Specifically,  the experimenter would apply correlation screening to a
sample of  concatenated vectors containing both  independent variables and  response variables.  Any  independent variable discoveries that have high cross-correlation with a response variable would be excellent candidates to include in the regression algorithm.

The outline of the paper is as follows. In Section \ref{sec:prelim}
the main assumptions are stated and the mathematical notation is
given. In Section \ref{sec:correlationscreening} the different kinds
of correlation screening tests are defined and the asymptotic theory
is developed and discussed. In section \ref{sec:sparse} the asymptotic
theory is specialized to the case of block-sparse population covariance. In
Section \ref{sec:applications} numerical results and experiments are
presented to illustrate the theory. Proofs of the principal results in
the paper are given in the Appendix/Supplemental Section. We also refer the reader to a technical report which contains more details on the results in this paper (see \cite{HeroRajaratnam:10}).

\section{Preliminaries}
\label{sec:prelim}

\def\mU{{\mathbb U}}
\def\mZ{{\mathbb Z}}
\def\mY{{\mathbb Y}}
\def\mX{{\mathbb X}}
In this section we set the notation and
recall some classical results on sample correlation.  See Anderson \cite{Anderson:03},  for example,
for more background.

 Let $\bX=[X_{1},\ldots, X_{p}]^T$ be a vector of random variables with mean
 $\boldsymbol{\mu}$ and $p\times p$ covariance matrix $\mathbf
 \Sigma$. Define the correlation matrix $\mathbf \Gamma = \mathbf
 D_\Sigma^{-1/2} \mathbf \Sigma \mathbf D_\Sigma^{-1/2}$ where
 $\bD_\Sigma= {\mathrm {diag}}_i (\mathbf \Sigma_{ii})$ is the
 diagonal matrix of variances of components of $\bX$. Assume that $n$
samples of $\bX$ are
available and arrange these samples in a $n\times p$ {\it data
   matrix}
$$\mathbb X = [\bX_1,\cdots, \bX_p]=[\bX_{(1)}^T, \cdots,
  \bX_{(n)}^T]^T,$$ where $\bX_i=[X_{1i}, \ldots, X_{ni}]^T$ and
$\bX_{(i)}=[X_{i1}, \ldots, X_{ip}]$ denote the $i$-th column and row,
respectively, of $\mathbb X$. Note that most of the results in this paper
hold when the rows of $\mathbb X$ are dependent.

Define the sample mean of the $i$-th column $\ol{X}_i=n^{-1} \sum_{j=1}^n X_{ji}$,
the vector of sample means $\ol{\bX}=[\ol{X}_1,\ldots, \ol{X}_p]$, the
$p\times p$ sample covariance matrix $\bS=\frac{1}{n-1} \sum_{i=1}^n
(\bX_{(i)}-\ol{\bX})^T(\bX_{(i)}-\ol{\bX})$, and the $p \times p$ sample
correlation matrix $\bR=\bD_{\bS}^{-1/2}\bS\bD_{\bS}^{-1/2}$, where
$\bD_{\bS}={\mathrm {diag}}_i (\bS_{ii})$ is the diagonal matrix of
component sample variances. Let the $ij$-th entry of the ensemble
covariance $\mathbf\Gamma$ be denoted $\gamma_{ij}$ and the $ij$-th
entry of the sample covariance $\bR$ be $\r_{ij}$.

The multivariate Z-scores $\bfZ_i \in \Reals^n$ are constructed by
standardizing the columns $\bX_i$ of $\mathbb X$ to have sample mean
equal to zero and sample variance equal to one
$$\bfZ_i= \frac{\bfX_i-\ol{X}_i \mathbf 1}{\sqrt{\bfS_{ii}(n-1)}} ,
\;\; i=1,\ldots, p,$$
where $\mathbf 1$ is a vector of ones. Equivalently,
$ \mathbb Z=[\bfZ_1, \ldots,\bfZ_p]=(n-1)^{-1/2} (\bI- n^{-1} \mathbf 1 \mathbf 1^T) \mathbb X \bD^{-1/2} .$
The Z-scores lie on the intersection of the $n-1$ dimensional
hyperplane $\{\bu \in \Reals^n: \mathbf1^T \bu =0\}$ and the
$n-1$ dimensional sphere $\{\bu \in \Reals^n: \|\bu\|_2 =1\}$.
The correlation matrix has the Z-score representation
$\bR=\mathbb Z^T\mathbb Z .$

An equivalent representation for the sample
correlation matrix $\bR$ uses what we call the U-scores, $\bU_i\in
\Reals^{n-1}$:
\be
\bR=\mathbb U^T\mathbb U ,\label{eq:Uscore_rep}
\ee
where $\mathbb U=[\bfU_1,
  \ldots,\bfU_p]$ is $(n-1)\times p$.  The U-scores lie on the $(n-2)$-sphere $S_{n-2}$ in
$\Reals^{n-1}$ and are constructed by projecting away the components
of the $\bX_i$'s orthogonal to the $n-1$ dimensional hyperplane $\{\bfu \in
\Reals^n: \mathbf1^T \bfu =0\}$, $i=1, \ldots, p$.
\def\bHn{{\mathbf H_{2:n}}} Specifically, define the orthogonal $n \times n$ matrix
$\mathbf H=[n^{-1/2}\mathbf 1, \mathbf H_{2:n}]$. The matrix $\bHn$
can be obtained by Gramm-Schmidt orthogonalization and satisfies the properties
$${\mathbf 1}^T
\mathbf H_{2:n} =[0,\ldots, 0], \;\; \bHn^T\bHn= \bI_{n-1}. $$
The U-score matrix $\mU=[\bU_1,\ldots,\bU_p]$ is obtained from $\mZ$ by the following relation
\be
\mU = \bHn^T \mZ.
\label{eq:Udef}
\ee

Furthermore, the sample correlation between $\bX_i$ and $\bX_j$ can be computed
using the inner product or the Euclidean distance between associated U-scores
\be
\r_{ij}= \bfU_i^T \bfU_j=1- \frac{\|\bU_i -\bU_j\|_2^2}{2}.
\label{eq:Uscore_dist_rep}
\ee

\begin{figure}
\begin{center}
\includegraphics[height=5cm]{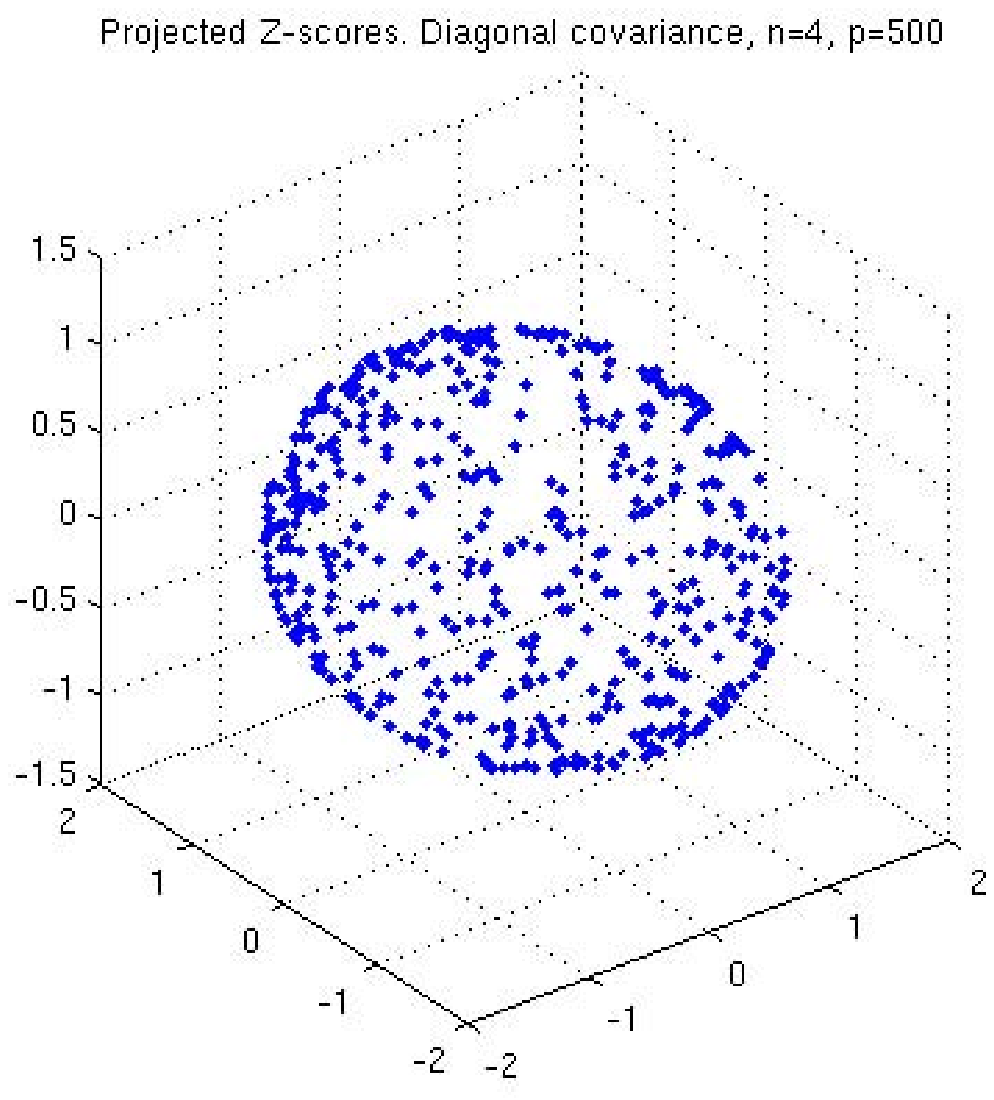}\;\;\;
\includegraphics[height=5cm]{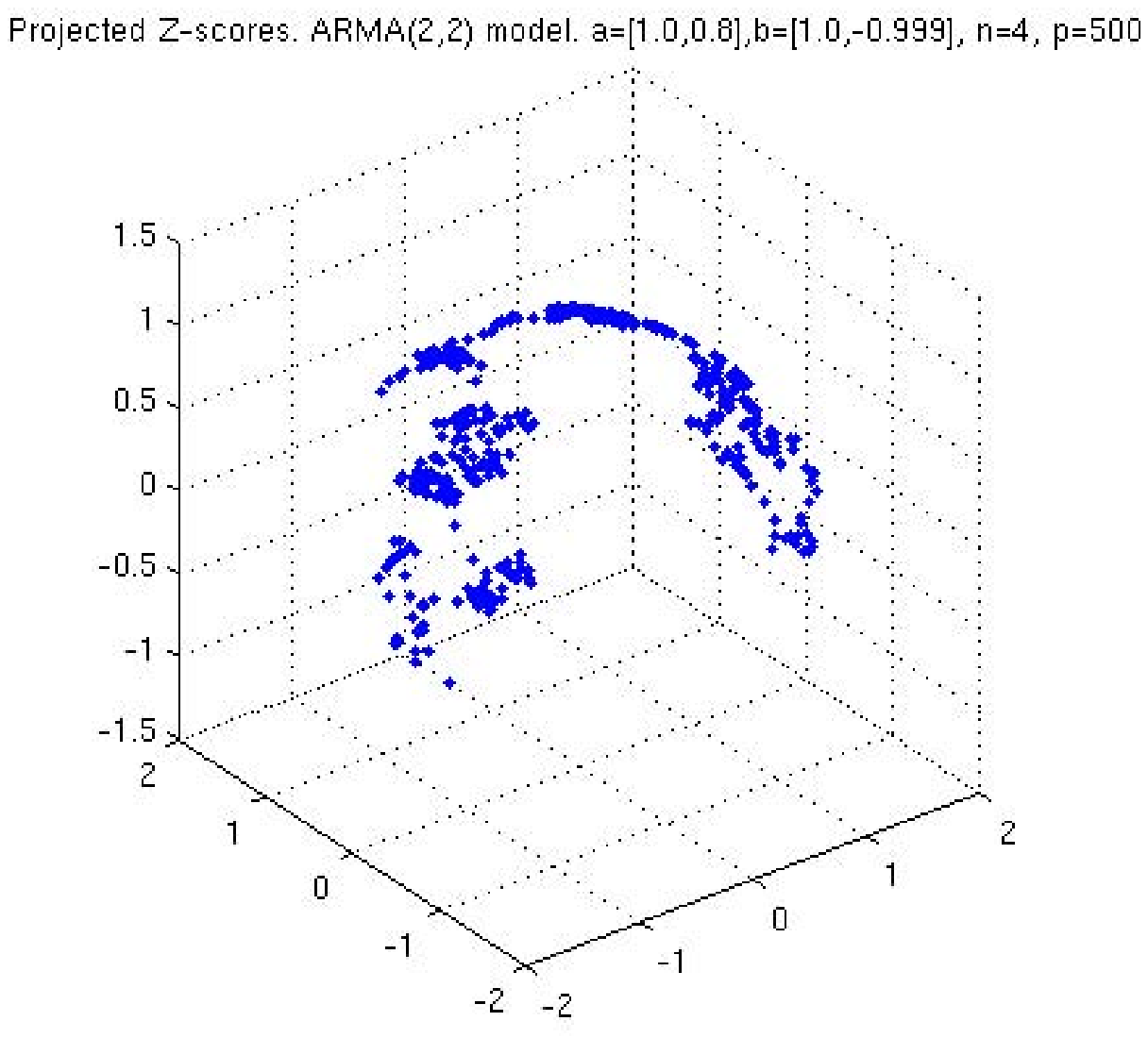}\\
\end{center}
\caption{The U-scores associated with $n=4$ realizations of 500
  variables are $n-1$-element vectors that lie on the unit $n-2$
  dimensional sphere $S_{n-2}$.  Shown are U-scores for a multivariate
  normal sample.  At left: for diagonal covariance matrix the 500
  U-scores are uniformly distributed over $S_{n-2}$. At right: for a
  non-diagonal covariance the U-scores are far from uniformly
  distributed on $S_{n-2}$. Pairs of U-scores that are close to each
  other, as measured by Euclidean distance, have high associated
  sample correlations. }
\label{fig:Uscores}
\end{figure}

As the U-score is an $(n-1)$-element vector it is a more compact
representation of the sample correlation than the $n$-element Z-score
vector. More importantly, the U-score lives in a geometry, the
$(n-2)$-sphere of co-dimension $1$ shown in Fig. \ref{fig:Uscores},
that is simpler than that of the standard Z-score.  

{\noindent \bf Elliptically contoured distributions}

The results in this paper hold for a wide class of sample
distributions that include light and heavy tailed distributions such
as the multivariate normal and multivariate student-t,
respectively. A random vector $\bX$ is said to follow an elliptical
distribution with location parameter $\boldsymbol{\mu}$ and dispersion matrix
parameter $\mathbf \Sigma$ if its density has the form \be
f_{\bX}(\bx) = |\mathbf \Sigma|^{-1/2}g\left((\bx - \boldsymbol{\mu})^T
\mathbf \Sigma^{-1} (\bx - \boldsymbol{\mu})\right) \label{eq:elliptical} ,
\ee
where $g(u)$ is a non-negative monotonic function.  When $\mathbf \Sigma$ is a diagonal matrix
the elliptical distribution is called {\it diagonal elliptical}.  It
is well known that when the rows of the data matrix $\mathbb X$ are
i.i.d. and follow a diagonal elliptical distribution the
U-scores are uniformly distributed on $S_{n-2}$, see for example
[Sec. 2.7]\cite{Anderson:03}. 
In the case of non-diagonal
$\mathbf \Sigma$ the distribution of the U-scores over the sphere
$S_{n-2}$ will generally be far from uniform (Fig. \ref{fig:Uscores}).
The U-score representations (\ref{eq:Uscore_rep}) and
(\ref{eq:Uscore_dist_rep}) of the sample correlation will be a key
ingredient for deriving the asymptotic results in this paper.

Invoked in the sequel will be the following sparsity
condition on the dispersion matrix.  The  matrix $\mathbf
\Sigma=((\sigma_{ij}))_{i,j}$  is said to be row-sparse of degree $k$  if
every row has fewer than  $k+1$  non-zero entries. Formally,
\be
  \{i:|\{j:\sigma_{ij}\neq 0\}| >k\} = \emptyset
\label{eq:rowsparse},
\ee
where $\emptyset$ is the empty set.
When the matrix is row-sparse of degree $q$
and there exists a permutation that block diagonalizes $\mathbf
\Sigma$ then the matrix satisfies the $q$-sparse condition of
Sec. \ref{sec:sparse}.

\vspace{0.3cm}
{\noindent \bf Relevant definitions:} The asymptotic expressions for the mean number of discoveries in the
next section will be a function of several quantities introduced below.

\vspace{0.3cm}

{\noindent \bf Spherical Cap Probability}

Define
  \be P_0=P_0(\rho,n)= \ra_n
\int_\rho^1\left(1-u^2\right)^{\frac{n-4}{2} } du,
\label{eq:Podef}
\ee
where $\ra_n$ is 
\be
\ra_n=\frac{2\Gamma((n-1)/2)} {\sqrt{\pi}
    \Gamma((n-2)/2)}
    \label{eq:andef}.
    \ee
The quantity $P_0/2$ is equal to the proportional area of the {\it
  spherical cap} of radius $r=\sqrt{2(1-\rho)}$ on
$S_{n-2}$.  It is the probability that a uniformly distributed point
$\bU$ on the sphere lies in pair of hyperspherical cones symmetric
about the origin. This probability expression was derived in the context of
the spherical normal distribution by Ruben
\cite[Eq. 4.1]{Ruben:AnnMathStat60}.
\footnote{The integral in
  \cite[Eq. 4.1]{Ruben:AnnMathStat60} is obtained from the integral in
  (\ref{eq:Podef}) by making change of variable $\theta=\arccos(u)$. }
A power series expansion of the integral in
(\ref{eq:Podef}) yields the relation, accurate as $\rho^2 $ approaches $1$:
\be 
P_0(\rho,n)= (n-2)^{-1} \ra_n (1-\rho^2)^{(n-2)/2}(1+O(1-\rho^2)) .
\label{eq:Porhorelation}
\ee

{\noindent \bf Relevant entropy and divergence quantities}

For a given density $f$ on  $S_{n-2}$ define the following entropy-related
functional, which satisfies the indicated inequality
\be H_2(f) = |S_{n-2}| \int_{S_{n-2}}f^2(\bfu) d\bfu
\geq 1
\label{eq:Renyi}.
\ee
Equality is attained in the inequality (\ref{eq:Renyi}) if and only if
(iff) $f$ is the uniform density: $f(\bfu)=|S_{n -2}|^{-1}$.  $H_2(f)$
is a monotonic transformation of the R\'enyi entropy of $f$ of order
$2$: $-\log \left(|S_{n-2}|^{-1}H_2(f)\right)$.

%
%

For a joint density $f_{\bfU,\bfV}$ on $S_{n-2}\times
S_{n-2}$ with marginals $f_\bfU$ and $f_\bfV$ define
\be J(f_{\bfU,\bfV})= |S_{n-2}|\int_{S_{n-2}} f_{\bfU,\bfV}(\bfu,\bfu) d\bfu .
\label{eq:biventropy}
\ee It will be shown that $J(f_{\bfU,\bfV})$ influences the mean
number of discoveries. Therefore, several intuitive
interpretations are given below that will be of use in the sequel.

First, $J(f_{\bfU,\bfV})$ is a measure of dependence between $\bU,
\bV$. Specifically, it is equal to the Bhattacharyya affinity
between $f_{\bfV}(\bfw)f_{\bfU}(\bfw)$ and the product
$f_{\bfU|\bfV}(\bfw|\bfw) f_{\bfV|\bfU}(\bfw|\bfw)$:
\be J(f_{\bfU,\bfV}) &=& |S_{n-2}| \int \sqrt{f_{\bfU|\bfV}(\bfw|\bfw)
  f_{\bfV|\bfU}(\bfw|\bfw)}\sqrt{ f_{\bfU}(\bfw) f_{\bfV}(\bfw)} d\bfw .
\label{eq:Bhattacharyya}
\ee
This is maximized when $\bU,\bV$ are statistically independent.

Second, the following asymptotic representation follows from (\ref{eq:thetafirstmomi}):
$$
J(\half f_{\bfU,\bfV}+\half f_{\bfU,-\bfV}) = \lim_{\rho\rightarrow 1} \frac{ P\left(\min\left\{\|\bfU-\bfV\|_2,\|\bfU+\bfV\|_2\right\}\leq \sqrt{2(1-\rho)}\right)}{P_0(\rho,n)}.
$$
%
The limit is equal to one when 
$\bfU$ and $\bfV$ are
independent and uniformly distributed on $S_{n-2}$.  Thus
$J(f_{\bfU,\bfV})-1$ is a measure of the deviation of the joint
density from uniform $f_{\bfU,\bfV}(\bfu,\bfv)=|S_{n-2}|^2$.  This
measure can either be positive, e.g., when $\bfU$ and $\bfV$ are
highly correlated or anti-correlated, or negative, e.g., when
$f_{\bfU,\bfV}(\bfu,\bfv)$ has nearly zero mass in the vicinity of the
diagonal $\bfu-\bfv =0$ and antidiagonal $\bfu+\bfv=0$
regions.

Finally, the following simple inequalities give further insight into
$J(f_{\bfU,\bfV})$:
\be J(f_{\bfU,\bfV}) &\leq& |S_{n-2}|\left(\int f_{\bfU|\bfV}(\bfw|\bfw)
f_{\bfV|\bfU}(\bfw|\bfw)d\bfw\right)^{1/2}\left( \int f_{\bfU}(\bfw)
f_{\bfV}(\bfw) d\bfw \right)^{1/2}\nonumber \\ &\leq&
H_2^{1/4}(f_{\bfU|\bfV})H^{1/4}_2(f_{\bfV|\bfU})H^{1/4}_2(f_{\bfU})H^{1/4}_2(f_{\bfV}),\label{eq:biventropybnd}
\ee
where equality in the first inequality and the second inequality
occur iff $f_{\bfU,\bfV}(\bfu,\bfu)=f_{\bfU}(\bfu)f_{\bfV}(\bfu)$ and
$f_{\bfU}(\bfu)=f_{\bfV}(\bfu)$, respectively. Hence $J(f_{\bfU,\bfV})$
is maximized when $\bfU$ and $\bfV$ are independent. In the other direction,
when restricted to the case of independent $\bfU$ and $\bfV$,
$J(f_{\bfU,\bfV})=H_2^{1/2}(f_{\bfU})H_2^{1/2}(f_{\bfV})$ is minimized when $\bfU$ and $\bfV$  are uniform over $S_{n-2}$.

\section{Correlation screening}
\label{sec:correlationscreening}

Consider an experiment to compare $p$ variables under treatments $a$ and $b$, called $\bfX^a$ and $\bfX^b$. The number $n$ of sample realizations may be different in the two
experiments but the number and identity of the $p$ variables are the
same. These experiments produce two data matrices: $\mathbb X^a $ and
$\mathbb X^b $, which are $n_a \times p$ and $n_b \times p$ matrices,
respectively.  From these data matrices extract the U-score matrices
$\mathbb U^a$ and $\mathbb U^b$.  Then, using the representation
(\ref{eq:Uscore_rep}), we construct
$\bR^a= [\mathbb U^a]^T \mathbb U^a$ and $\bR^b= [\mathbb U^b]^T \mathbb U^b$, and call them sample auto-correlation matrices. When $n_a=n_b$ we can also construct the
sample cross-correlation matrix $\bR^{ab}=[\mathbb U^a]^T \mathbb U^b$. We are primarily interested in the case $n_a,n_b \ll p$ so that the auto-correlation and cross-correlation matrices will be rank deficient. Let the
$ij$-th element of each of these matrices be denoted as $\r_{ij}^a$,
$\r_{ij}^b$, and $\r_{ij}^{ab}$, respectively.

We distinguish between three types of correlation screening. We use the terms auto-correlation and cross-correlation in analogy to auto-correlation and cross-correlation functions in time series analysis.

\noindent{\bf Auto-correlation screening}: The objective is to screen
the $p$ variables for those whose maximal magnitude correlation
exceeds a given threshold $\rho_a$. Specifically, for $i,j =1, \ldots,
p$, the $i$-th variable passes the screen if:
\be
\max_{j\neq
  i}|\r_{ij}^a|> \rho_a.
\label{eq:autoscreen}
\ee

\noindent{\bf Cross-correlation screening}: The objective is to screen
the $p$ variables for those whose maximal magnitude cross-correlation
exceeds a given threshold $\rho_{ab}$. Specifically, for $i,j =1,
\ldots, p$, the $i$-th variable passes the screen if:
\be
\max_{j\neq
  i}|\r_{ij}^{ab}|> \rho_{ab}.
\label{eq:crossscreen}
\ee

\noindent{\bf Persistent auto-correlation screening}: The objective is
to screen the $p$ variables for those whose maximal magnitude
auto-correlation in both treatments exceeds given thresholds
$\rho_{a}$ and $\rho_{b}$, respectively. Specifically, for $i =1,
\ldots, p$, the $i$-th variable passes the screen if:
\be
\max_{j\neq i}|\r_{ij}^{a}|> \rho_{a} \;\mbox{and}\;\max_{j\neq
  i}|\r_{ij}^{b}|> \rho_{b}.
\label{eq:persistentscreen}
\ee

\begin{figure}
\begin{center}
\includegraphics[height=5cm]{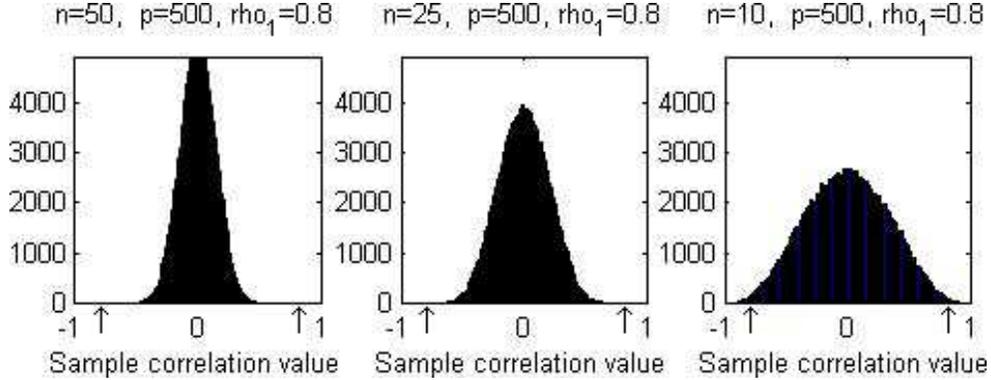}
\end{center}
\caption{Effect of number of samples $n$ on the
  discoveries for a multivariate normal  sample where all but
  two of the $p=500$ variables are mutually correlated as $n$ decreases over the range $50,25,10$. These two variables have a correlation coefficient equal to $\rho_1=0.8$.  Shown are histograms of the $p(p-1)/2$ distinct sample correlation coefficients in the correlation matrix $\bR$ excluding the diagonal coefficients.  The arrows point to the locations of the positive and negative correlation thresholds 
  of an auto-correlation screening test that would detect the variables having at least $0.8$ correlation with probability not exceeding $0.5$. An increasing  number of other sample correlations exceed this threshold as $n$ decreases: these false discoveries are overwhelming for small $n$.}
\label{fig:phasetransition}
\end{figure}

\def\ab{{a\wedge b}}
For each of the above three tests a discovery is declared if an index
$i$ passes the screen and we denote by $N^a$, $N^{ab}$, and
$N^{\ab}$, respectively, the total number of discoveries.  For large
$p$, these three tests display similar phase transition phenomena.
For example, we illustrate in Fig. \ref{fig:phasetransition} how the
number $N^a$ of false auto-correlation discoveries experiences a
sharp increase as the threshold $\rho_a$ is reduced beyond a certain
critical value $\rho_c$. This critical value depends on the number $p$
of variables, the number $n=n_a$ of  samples, and the joint
distribution of the $p$ variables. The behavior gets worse as $n$
decreases relative to $p$, eventually overwhelming the test with false
discoveries for all but a narrow range of thresholds $\rho$ close to $1$.

In the next three subsections we develop theory to predict this phase
transition behavior in terms of the mean number of discoveries.

\subsection{Discoveries in auto-correlation screening}
\label{sec:autoscreening}

Here we give results for the mean number of discoveries $E[N^a]$ when
screening for threshold-exceeding correlations between columns of a
single data matrix $\mathbb X^a$. For convenience here we suppress
``$a$" superscripts and subscripts.

We recall the quantities 
\be
\gamma_{p}=\max_{1\leq k<p}\{a_n^k M_{k|1}\}, \;\;   \eta_p=2 a_n^2 \dot{M}_{1|1}
\label{eq:gammaeta}
\ee where $a_n= |S_{n-2}|$, $M_{k|1} $ is defined in (\ref{eq:Mk1def})
and $\dot{M}_{1|1}$ is defined in (\ref{eq:mdotdef}). These quantities are uniformly bounded over $p$ when the joint density $f_{\bfU_1, \ldots, \bfU_p}$ of the U-scores is smooth and strictly bounded between $(0,\infty)$.
For example if the joint density of the $Z$-scores is a finite mixture of von Mises-Fisher densities on the sphere $S_{n-2}$ with strictly bounded concentration parameters, then $\gamma_{p}$ and  $\eta_p$ are uniformly bounded.

%
%

\begin{propositions}
Let the $n \times p$ data matrix $\mathbb X$ have associated U-scores
$\mathbb U$ and assume that $n>2$.  Assume that $\gamma_p$ and
$\eta_p$ are uniformly bounded. Let the sequence $\{\rho_p\}_p$ of
correlation thresholds be such that $\rho_p\rightarrow 1$ and
$p(p-1)\left(1-\rho_p^2\right)^{(n-2)/2} \rightarrow e_n$ for some
finite constant $e_n$. Then the mean number of discoveries generated
from the auto-correlation screen (\ref{eq:autoscreen}) satisfies: \be
\left|E[N]- \kappa_n
J(\ol{f_{\bfU_{\bullet},\bfU_{\ast-\bullet}}})\right| \leq O(p^{-1})+
O(\sqrt{1-\rho_p}),
\label{eq:ENdef}
\ee
where $\kappa_n =\ra_n e_n/(n-2)$ and
\be \ol{f_{\bfU_{\bullet},\bfU_{\ast-\bullet}}}(\bfu,\bfv)=\frac{1}{p}\sum_{i=1}^p \frac{1}{p-1}\sum_{j\neq i}^p \left(
\half f_{\bfU_i,\bfU_j}(\bfu,\bfv)+\half
f_{\bfU_i,\bfU_j}(\bfu,-\bfv)\right),
\label{eq:avgfubiv}
\ee is the average of the pairwise U-score density.  Assume in
addition that the joint density of the U-scores satisfies the weak
dependency condition: for some $k=o(p)$ the average dependency
coefficient $\|\Delta_{p,k}\|_1$  (\ref{eq:Deltapdefavg})
converges to zero.  Then $P(N>0) \rightarrow 1-\exp(-\Lambda/2)$ where
$\Lambda$ is the limiting value of $E[N]$ specified by
(\ref{eq:ENdef}).
\label{prop1}
\end{propositions}

In the proof of Prop. \ref{prop1} we establish the stated limit on
$P(N>0)$ by showing that $N$ is dominated by the number $N_e$ of edges
in the correlation graph and that $N_e$ converges to a Poisson random
variable $N^*$ with rate $\Lambda/2$ as $p \rightarrow \infty$. The
rate of convergence of $P(N>0)$ to the stated limit
is of order $\max\{(k/p)^2, \|\Delta_{p,k}\|_1\}$.

 In terms of the limiting value (\ref{eq:ENdef}) of
  $E[N]$ the case where the columns of $\mathbb X$ have spherically
  contoured distribution  is of special interest. In this case the
  U-scores are i.i.d. uniformly distributed and 
  $J(\ol{f_{\bfU_{\bullet},\bfU_{\ast-\bullet}}})=1$.  
  Prop. \ref{prop1} asserts the weaker necessary and sufficient condition:
  $J(\ol{f_{\bfU_{\bullet},\bfU_{\ast-\bullet}}})=1$ if and only if
  the averaged pairwise U-score density (\ref{eq:avgfubiv}) is
  i.i.d. uniform over $S_{n-2}\times S_{n-2}$. We develop this
further in the next paragraph.

First observe that the marginal densities, obtained by integrating
$\ol{f_{\bfU_{\bullet},\bfU_{\ast-\bullet}}}(\bfu,\bfv)$ over $\bfv$ and
$\bfu$, are identical and equal to the average U-score density
\be \ol{f_{\bfU_\ast}}(\bfu)= \frac{1}{p} \sum_{i=1}^p \left(\half
f_{\bfU_i}(\bfu)+\half f_{\bfU_i}(-\bfu)\right).
\label{eq:avgfub}
\ee
Therefore inequality (\ref{eq:biventropybnd}) implies that
\be
J(\ol{f_{\bfU_{\bullet},\bfU_{\ast-\bullet}}}) \leq H_2^{1/4}(f_{U|V})H_2^{1/4}(f_{V|U}) 
H_2^{1/2}(\ol{f_{\bfU_\ast}}),
\label{eq:extremalJ}
\ee
with equality iff
$\ol{f_{\bfU_{\bullet},\bfU_{\ast-\bullet}}}(\bfu,\bfu)=
(\ol{f_{\bfU_\ast}}(\bfu))^2$, which satisfied when the U-scores are 
independent. 
Second observe that the extremal property (\ref{eq:Renyi}) of $H_2(f)$
implies that, among all such i.i.d. U-score distributions, $E[N]$ will
be smallest when the marginal $\ol{f_{\bfU_\ast}}$ is uniform, which is satisfied when the U-scores are uniform on $S_{n-2}$.

In the case that
$\ol{f_{\bfU_{\ast},\bfU_{\ast-\bullet}}}(\bfu,\bfu)=|S_{n-2}|^{-2}$, (\ref{eq:ENdef}) implies
the asymptotic approximation for finite $p$ and $\rho<1$: 
\be
E[N]\approx\kappa_n\approx p(p-1) P_0(\rho,n) \label{eq:ENdefunif} ,
\ee
since $p(p-1) P_0(\rho_p,n) \rightarrow \kappa_n$ as $p \rightarrow \infty$. 
This case holds, for example, when  the rows of $\mathbb X$ 
are i.i.d. with diagonal elliptical distribution. In this case
the U-scores are i.i.d. uniform and the mean number of discoveries has
the exact expression
\be E[N]=p(1-(1-P_0(\rho,n))^{p-1}) \label{eq:ENdefunife}.\ee
%
%

\begin{figure}
\begin{center}
\includegraphics[height=5cm]{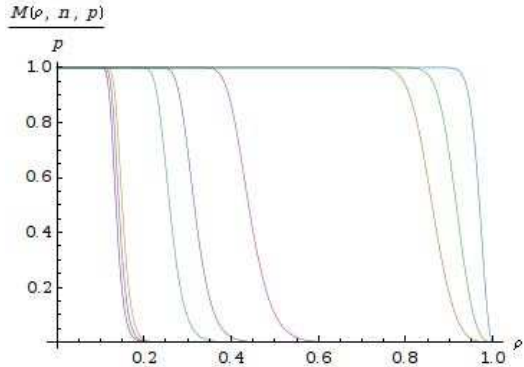}\\
\end{center}
\caption{Normalized mean number of discoveries $E[N]/p$ for the case
  that the rows of the data matrix are normal with diagonal
  covariance.  Nine curves are plotted as a function of the screening
  threshold $\rho$ for $p=500$ and nine values of $n$. The values
  $n=550, 500, 450, 150, 100, 50, 10, 8, 6$ index the curves from left
  to right.}
\label{fig4}
\end{figure}

\begin{table}
\begin{center}
\begin{tabular}{|c||c|c|c|c|c|c|c|c|c|c|}  \hline
n & 550 & 500 & 450 & 150 & 100 & 50 & 10 & 8 & 6 \\  \hline
$\rho_c$ &0.188&0.197&0.207&0.344&0.413&0.559&0.961&0.988&0.9997 \\ \hline
\end{tabular}
\end{center}
\caption{Values of the critical threshold $\rho_c$ where phase
  transition occurs in Fig. \ref{fig4}. These values were determined
  using asymptotic approximation (\ref{eq:rhocrit}).}
\label{table1}
\end{table}

In Fig. \ref{fig4} we plot the exact expression (\ref{eq:ENdefunife})
for the normalized mean number of discoveries as a function of $\rho$
and $n$ for $p=500$.  Each curve, decreasing monotonically as $\rho$
increases, is a plot of $E[N]/p$ for given $n$. Since the true
covariance matrix is diagonal all discoveries are false discoveries. 
We make several observations:
\begin{itemize}
\item
The curves in Fig. \ref{fig4} cluster into three groups. From left to
right: $n\in \{550,500,450\}$, $n\in \{150,100,50\}$ and $n\in
\{10,8,6\}$.  The effect on the curves of varying $n$ is more
pronounced for small $n$  than for larger $n$.
\item
The curves illustrate a phase transition phenomenon in the mean number
of false positives as a function of the threshold $\rho$. For given
$n$ there is a critical point $\rho_c$ such that as $\rho$ approaches
$\rho_c$ from above the mean number of false positives is small and
increases very slowly. As $\rho$ continuous to decrease in the
vicinity of $\rho_c$ the mean number of false positives increases
rapidly to $p$.
\item
The rapidity of the phase transition varies as a function of $n$ and is related to the slope of the curve 
near its inflection point. The most rapid phase transitions occur when
$n$ is very large or very small.
\end{itemize}

The phase transition threshold value $\rho_c$ can be predicted by the
knee of the curve in Fig. \ref{fig4}, defined as the maximum value
$\rho$ at which the slope of the curve equals minus one.  This choice of critical slope is common in the physics literature. One could choose
a different critical slope value to define $\rho_c$ but this would only have a minor effect (a change in the quantity $c_n$ in (3.10) by a constant scale factor).
The slope of the large $p$ approximation (\ref{eq:ENdef}) to $E[N]$ is
$$
dE[N]/d\rho=-p(p-1)(1-\rho^2)^{(n-4)/2} \ra_n J(\ol{f_{\bfU_{\bullet},\bfU_{\ast-\bullet}}}),
$$
where  $\ra_n$ is given in (\ref{eq:andef}).
Define the critical value as $\rho_c=\max\{\rho:
p^{-1}dE[N]/d\rho=-1\}$.  For $n>4$ this is maximization can
be solved to give the  expression
%
%
\be
\rho_c=\sqrt{1-c_n(p-1)^{-2/(n-4)}},
\label{eq:rhocrit} \ee
where
$c_n=\left(\ra_nJ(\ol{f_{\bfU_{\bullet},\bfU_{\ast-\bullet}}})\right)^{-2/(n-4)}$.  
The
accuracy of $\rho_c$ defined in (\ref{eq:rhocrit}) can be appreciated
by comparing the predicted $\rho_c$ in Table \ref{table1} to the
transition points of the associated curves in Fig. \ref{fig4}.

\subsection{Discoveries in cross-correlation screening}
\label{sec:crossscreening}

Next we turn to screening for threshold-exceeding cross-correlations
between columns of two data matrices $\mathbb X^a$ and $ \mathbb X^b$.
The theory in the previous section could be directly used by applying
Prop. \ref{prop1} to the concatenated $ n\times 2p$ data matrix
$$\mathbb X=\left[\begin{array}{cc} \mathbb X^a & \mathbb
    X^b\end{array} \right].$$
However, the convergence rates and phase transition thresholds would
be significantly worse than before due to the inflation of the number
of variables from $p$ to $2p$. Furthermore, if we thresholded the
entire $2p \times 2p$ sample correlation matrix $\mathbb X^T \mathbb
X$ we would expect that in most practical problems the
auto-correlation discoveries in the diagonal blocks would dominate the
cross-correlation discoveries in the off-diagonal blocks. The following
result is useful when one is only interested in the cross-correlation
discoveries.

Define $\gamma_p^{ab}$ and $\eta_p^{ab}$ similarly to
(\ref{eq:gammaeta}) except that $M_{k|1}$ and $\dot{M}_{1|1}$ are replaced 
by $M_{k|1}^{ab}$ and $\dot{M}_{1|1}^{ab}$ as defined in (\ref{eq:Mk1abdef}) and 
(\ref{eq:mdotabdef}).
 
\begin{propositions}
Let the $n \times p$ data matrices $\mathbb X^a$ and $\mathbb X^b$
have associated U-scores $\mathbb U^a$ and $\mathbb U^b$ and assume that
$n>2$.  Assume that $\gamma_p^{ab}$ and $\eta_p^{ab}$ are uniformly bounded. 
Let the sequence $\{\rho_p\}_p$ of cross-correlation thresholds be
such that $\rho_p\rightarrow 1$ and
$p^2\left(1-\rho_p^2\right)^{(n-2)/2} \rightarrow e_n$
for some finite constant $e_n$. Then the mean number of discoveries generated from the cross-correlation screen (\ref{eq:crossscreen}) satisfies:
\be
\left| E[N^{ab}] - \kappa_n J(\ol{f_{\bfU^a_{\ast},\bfU^b_{\bullet}}}) \right|\leq
O(p^{-1})+ O(\sqrt{1-\rho_p}),
\label{eq:ENdefcross}
\ee
where $\kappa_n =\ra_n e_n/(n-2)$ and 
\be \ol{f_{\bfU^a_{\ast},\bfU^b_{\bullet}}}=
\frac{1}{p}\sum_{i=1}^p \frac{1}{p}\sum_{j=1}^p \left( \half
f_{\bfU^a_i,\bfU^b_j}(\bfu,\bfv)+\half
f_{\bfU^a_i,\bfU^b_j}(\bfu,-\bfv)\right) .
\label{eq:avgfubivcross}
\ee Assume in addition that the joint density of the U-scores
satisfies the weak cross-dependency condition: for some $k=o(p)$ the
average dependency coefficient $\|\Delta_{p,k}^{ab}\|_1$
 (\ref{eq:Deltapdefavg}) converges to zero.  Then $P(N^{ab}>0)
\rightarrow 1-\exp(-\Lambda)$ where $\Lambda$ is the limiting value of 
$E[N^{ab}]$ specified by
(\ref{eq:ENdefcross}).
\label{prop1cross}
\end{propositions}

The critical phase transition threshold for the case of
cross-correlation screening can be derived in a similar manner to the
previously considered case of auto-correlation screening. The critical
threshold is given by
\be \rho_c=\sqrt{1-c_n^{ab}p^{-2/(n-4)}},
\label{eq:rhocritcorrscreenpers} \ee
where $c_n^{ab}=\left(\ra_nJ(\ol{f_{\bfU^a_{\bullet},\bfU^b_{\bullet}}})\right)^{-2/(n-4)}$ and $\ra_n$ is given in (\ref{eq:andef}).

\subsection{Discoveries in persistent-correlation screening}
\label{sec:persistentcreening}

Finally we treat screening for variables whose auto-correlation
exceeds a threshold in both of two treatments $a$ and $b$. Recall that
in this problem there are two correlation thresholds $\rho^a$ and
$\rho^b$ that are respectively applied to the $p\times p$ sample
correlation matrices derived from the independent data matrices
$\mathbb X^a$ and $\mathbb X^b$.  As discussed below, Prop.
\ref{prop1} could be directly applied to this problem but it would
result in an uninteresting degenerate limit. A more interesting result
is the following.


\begin{propositions}
Let the $n_a \times p$ data matrix $\mathbb X^a$ and the $n_b\times p$
data matrix $\mathbb X^b$ be statistically independent and assume that
the associated U-scores from each treatment satisfy the same
conditions assumed for in Prop. \ref{prop1}.  Let the sequences
$\{\rho_p^a\}_p$ and $\{\rho_p^b\}_p$ be such that
$\rho_p^a\rightarrow 1$ and $\rho_p^b\rightarrow 1$ while
$p^{1/2}(p-1)\left(1-(\rho_p^a)^2\right)^{(n_a-2)/2} \rightarrow
e_{n_a}$ and $p^{1/2}(p-1)\left(1-(\rho_p^b)^2\right)^{(n_b-2)/2}
\rightarrow e_{n_b}$ for some finite constants $e_{n_a}, e_{n_b}$.
Then the mean number of discoveries $N^{\ab}$ generated by the
persistent-correlation screen (\ref{eq:persistentscreen}) satisfies
\be
\label{eq:ENdefpersistent}
\\
\left|E[N^{\ab}] - \kappa_n^{\ab}
 \frac{1}{p}\sum_{i=1}^p
J(\ol{f_{\bfU^a_{i},\bfU^a_{\ast-i}}})J(\ol{f_{\bfU^b_{i},\bfU^b_{\ast-i}}}) \right| 
\leq O\left(\max\{(k/p)^3,
(k/p)p^{-1/2}, p^{-1}, \|\Delta_{p,k}^a\|_1,\|\Delta_{p,k}^b\|_1\}\right),
\nonumber\ee
where 
$\kappa_n^{\ab} = e_{n_a} e_{n_b} \ra_{n_a}\ra_{n_b} (n_a-2)^{-1}(n_b-2)^{-1}$ 
and, for $\bfU \in \{\bfU^a,\bfU^b\}$, $\ol{f_{\bfU_{i},\bfU_{\ast-i}}}$
is the leave-one-out
average of the U-score pairwise densities:
\be \ol{f_{\bfU_{i},\bfU_{\ast-i}}}(\bfu,\bfv)=\frac{1}{p-1}\sum_{j\neq i}^{p} \left( \half f_{\bfU_i,\bfU_j}(\bfu,\bfv)+\half f_{\bfU_i,\bfU_j}(\bfu,-\bfv)\right).
\label{eq:avgfubivpersistent}
\ee
%
Assume in addition that the U-score densities associated with
$\mathbb X^a$ and $\mathbb X^b$ each satisfy the weak dependency
condition stated in Prop. \ref{prop1}.  Then $P(N^{\ab}>0) \rightarrow
1-\exp(-\Lambda)$ where $\Lambda$ is the limiting value of $E[N]$ specified in
(\ref{eq:ENdefpersistent}).
\label{prop1persistent}
\end{propositions}

In Prop. \ref{prop1persistent} the assumed rates
of convergence of $\rho_p^a,\rho_p^b$ are slower (note the different
factor $p^{1/2}$) than the rates assumed in Prop. \ref{prop1} and
\ref{prop1cross}. A slower rate is required since persistent
correlation discoveries are rarer than auto-correlation discoveries.
In particular, when the correlation thresholds satisfy the hypotheses
of Prop. \ref{prop1persistent} the individual per-treatment means
$E[N^a]$ and $E[N^b]$ do not converge. However, it can be shown that
$p^{-1/2}E[N^a]$ and $p^{-1/2}E[N^b]$ do converge (see Corollary
\ref{corrpersistent} in Appendix/Supplemental Section). Conversely, if the individual
per-treatment means converge to finite values then the mean number of
persistent discoveries $E[N^{\ab}]$ converges to zero, resulting in an
uninteresting limit.

Assume that one or the other of the factors in the summand of
(\ref{eq:ENdefpersistent}) do not depend on $i$:
\be
J(\ol{f_{\bfU^a_{i},\bfU^a_{\ast-i}}})= J(\ol{f_{\bfU^a_{\bullet},\bfU^a_{\ast-\bullet}}})
, \;\;\; {\mathrm{or}} \;\;
J(\ol{f_{\bfU^b_{i},\bfU^b_{\ast-i}}})=J(\ol{f_{\bfU^b_{\bullet},\bfU^b_{\ast-\bullet}}}).
\label{eq:Ja}\label{eq:Jb}
\ee
When (\ref{eq:Ja}) holds we say that the pairwise dependencies are
    {\it incoherent} across treatments $a$ and $b$. A sufficient condition for
incoherence is pairwise independent U-scores with identical marginal densities
$f_{\bfU_i}^a=f_{\bfU_j}^a$ and $f_{\bfU_i}^b=f_{\bfU_j}^b$. %
%
%
In the incoherent case the limit (\ref{eq:ENdefpersistent}) takes on a
simpler intuitive form
$$
\frac{1}{p}\sum_{i=1}^p
J(\ol{f_{\bfU^a_{i},\bfU^a_{\ast-i}}})J(\ol{f_{\bfU^b_{i},\bfU^b_{\ast-i}}})
= J(\ol{f_{\bfU^a_{\bullet},\bfU^a_{\ast-\bullet}}})J(\ol{f_{\bfU^b_{\bullet},\bfU^b_{\ast-\bullet}}}).
$$
Define $\kappa_{n_a} = p^{1/2}e_{n_a}\ra_{n_a}/(n_a-2)$ and $\kappa_{n_b} =
p^{1/2}e_{n_b}\ra_{n_b}/(n_b-2)$.  
Then, in view of the limit (\ref{eq:ENdef})
of Prop. \ref{prop1}, under the condition (\ref{eq:Ja}) the limit in
(\ref{eq:ENdefpersistent}) gives the large $p$ approximation
%
%
\be E[N^{\ab}] \approx \frac{E[N^a]E[N^b]}{p}
\label{eq:NaNb} .
\ee
The right side of (\ref{eq:NaNb}) is equal to the right side of
(\ref{eq:ENdefpersistent}) when the pairwise dependencies are
incoherent across treatments $a$ and $b$.


Relation (\ref{eq:NaNb}) is a well known asymptotic relation for the
number of matches in two independent Bernoulli sequences of length
$p$. In this case $N^{\ab}$ is the number of successes common to the
pair of sequences and $N^a$,$N^b$ are the number of successes in each
sequence; a result easily established using for large $p$ Stirling
approximations and assuming small probabilities of success.  It is
interesting that in persistency screening it is sufficient that only
one of the two treatments produce identically distributed U-scores for
(\ref{eq:NaNb}) to hold.

We next turn to the problem of selecting the thresholds $\rho^a$
and $\rho^a$. These thresholds affect the asymptotic mean number of
discoveries (\ref{eq:ENdefpersistent}) only through the limits $e_{n_a}$
and $e_{n_b}$ defined in Prop. \ref{prop1persistent}
When relation (\ref{eq:NaNb}) holds, it can be shown that if we fix
the normalized average rate of per-treatment discoveries
$(E[N^a]+E[N^b])p^{-1/2}/2$, $E[N^{ab}]$ is maximized when the
thresholds $\rho_a$ and $\rho_b$ are chosen to make $E[N^a]=E[N^b]$.
These optimal thresholds are related by
$$
1-\rho_a^2= (1-\rho_b^2)^{\frac{n_a-2}{n_b-2}}\left(\frac{(n_b-2)\ra_{n_a} J(\ol{f_{\bfU^a_\bullet,\bfU^a_{\ast-\bullet}}})}{(n_a-2)\ra_{n_b} J(\ol{f_{\bfU^b_\bullet,\bfU^b_{\ast-\bullet}}})}\right)^{2/(n_a-2)}.$$

A general closed form expression for the critical phase transition
threshold for persistent-correlation screening has not been
found. However, for the special case of pairwise i.i.d. U-scores and
equal number $n=n_a=n_b$ of samples, the following
expression for the critical threshold holds
\be
\rho_c=\sqrt{1-c_n^{\ab}(p-1)^{-2/(n-4)}},
\label{eq:rhocritcorrscreen} \ee
where $c_n^{\ab}=\left(\ra_n \left(H_2(\ol{f_{\bfU^a_\ast}})H_2(\ol{f_{\bfU^b_\ast}})\right)^{1/2}\right)^{-2/(n-4)}$ and $\ra_n$ is given in (\ref{eq:andef}).

Prop \ref{prop1persistent} generalizes to more than two
treatments. Assume there are $m$ different independent treatments
$t_1, \ldots , t_m$ then the correlation thresholds $\rho_p^{t_j}$
should be selected such that they converge to one and
$p^{1/m}(p-1)\left(1-(\rho_p^{t_j})^2\right)^{(n_{t_j}-2)/2}$
converges to a finite constant, say $e_{n_j}$, $j=1, \ldots, m$. In
this case one obtains the same type of limit of the false
positive rate as in Prop. \ref{prop1persistent} under similar
conditions of  weak dependence of the variables
within each treatment. The
mean number of discoveries will converge to
\be \lim_{p\rightarrow
  \infty}E[N^{t_1\wedge\cdots\wedge t_m}] = \kappa_n
\lim_{p\rightarrow \infty} p^{-1}\sum_{i=1}^p \prod_{j=1}^m
J(\ol{f_{\bfU^{t_j}_{i},\bfU^{t_j}_{\ast-i}}})
\label{eq:ENdefpersistentgen}
\ee
where $\kappa_n = \prod_{j=1}^m \left(\frac{\ra_{n_j}e_{n_j}}{n_j-2}\right)$.
If $\min\{n_j\}> 2(m+1)$ and $\dot{M}_{1|1}$ (defined in
(\ref{eq:mdotdef})) is bounded, the rate of convergence in
(\ref{eq:ENdefpersistentgen}) will be dominated by the treatment with
the fewest samples and it will be of order
$O\left(p^{-2/(n-2)}\right)$ where $n=\min_j\{n_j\}$. Otherwise the
rate of convergence will be $O(p^{-1/m})$. When the factors in the summand
of the limit (\ref{eq:ENdefpersistentgen}) do not depend on $i$ a
relation analogous to (\ref{eq:NaNb}) holds:
$E[N^{t_1\wedge\cdots\wedge t_m}]\approx (E[N^{t_1}]\cdots
E[N^{t_m}])/p^{m-1}$.

\section{Correlation screening with sparse dependency}
\label{sec:sparse}

In this section we specialize to the class of $q$-sparse $p\times p$
covariances, defined as row-sparse covariance matrices of
degree $q$ that can be reduced to a single $q\times q$ block of correlated
variables using row-colum permutations.
Under this $q$-sparse condition, to order
$O\left(\left(q/p\right)^2\right)$ the limits stated in Propositions
\ref{prop1}-\ref{prop1persistent} do not depend on the unknown joint
sample distribution. Therefore, these propositions can be used to
determine universal screening thresholds that approximately control
any desired level of false positive rate. We treat each of the three
correlation screening procedures separately.

\subsection{Sparse auto-correlation screening}

Let  the rows of $\mathbb X$ be i.i.d.
Under the assumption that the  columns  of $\mathbb X$ have
$q$-sparse covariance, the U-scores $\{\bU_i^a\}_{i=1}^p$
are i.i.d. uniform except for a number of $q \leq p$ mutually
dependent U-scores $\{\widetilde{\bU}_i^a\}_{i=1}^q$ that are
independent of the rest.
The  mean number of discoveries in
Prop. \ref{prop1} becomes, to order at most $O\left(\max\left\{p^{-1},p^{-2/(n-2)}\right\}\right)$,
$$
E[N^a]=\kappa_n^a\left(1+ \frac{q(q-1)}{p(p-1)} \left(J(\ol{f_{\widetilde{\bfU}_{\bullet}^a,\widetilde{\bfU}_{\ast-\bullet}^a}})-1\right)\right) ,
$$
where $\ol{f_{\widetilde{\bfU}_{\bullet}^a,\widetilde{\bfU}_{\ast-\bullet}^a}}$
is the average over the joint distributions of distinct and mutually
dependent U-scores.
Therefore, to order at most $O\left(\max\left\{(q/p)^2, p^{-1},p^{-2/(n-2)}\right\}\right)$ the mean number of discoveries is equal to $\kappa_n^a$.

\subsection{Sparse cross-correlation screening}

Let the rows of $\mathbb X$ be i.i.d.  Assume that the cross
correlation matrix is block-sparse in the sense that there exists a
column permutation that puts the cross-correlation matrix into a form
having most entries zero except for a small $q_{a}\times q_{b}$
non-zero off diagonal block.
Then  the mean number of discoveries in
Prop. \ref{prop1cross} becomes, to order at most $O\left(\max\left\{p^{-1},p^{-2/(n-2)}\right\}\right)$,
$$ E[N^{ab}]=\kappa_n^{ab}\left(1+ \frac{q_{a}q_{b}}{p^2}
\left(J(\ol{f_{\widetilde{\bfU}_{\bullet}^a,\widetilde{\bfU}_{\bullet}^b}})-1\right)\right) .
$$
Therefore, with $q=\max\{q_a,q_b\}$, to order $O\left(\max\left\{(q/p)^2, p^{-1},p^{-2/(n-2)}\right\}\right)$
the mean number of discoveries is equal to $\kappa_n^{ab}$.

\subsection{Sparse persistent-correlation screening}

Let  the rows of $\mathbb X$ be i.i.d.
Assume that under treatment $a$ all variables are mutually
uncorrelated except for a those variables with indices in the set
$Q_a$.  Likewise define the index set $Q_b$ of variables having
non-zero correlation under treatment $b$.  The  mean number of discoveries in Prop. \ref{prop1persistent} becomes, to order $O\left(\max\left\{p^{-1/2},p^{-2/(n-2)}\right\}\right)$,
\ben E[N^{\ab}]/\kappa_n^{\ab}&=&1+
\left(\frac{|Q_b-Q_a|}{p}\right)\left(\frac{|Q_b|-1}{p-1}
\right)(\tilde{J}_b-1) +
\left(\frac{|Q_a-Q_b|}{p}\right)\left(\frac{|Q_a|-1}{p-1}\right)
(\tilde{J}_a-1) \\ &&+ \left(\frac{|Q_a\cap
  Q_b|}{p}\right)\left(\frac{|Q_a|-1}{p-1}\right)
(\tilde{J}_a-1)+\left(\frac{|Q_a\cap
  Q_b|}{p}\right)\left(\frac{|Q_b|-1}{p-1}\right) (\tilde{J}_b-1)
\\ &&+\left(\frac{|Q_a\cap Q_b|}{p}\right)\left(\frac{|Q_a|-1}{p-1}\right)\left( \frac{|Q_b|-1}{p-1}\right)
(\tilde{J}_a-1) (\tilde{J}_b-1), \een
where $\tilde{J}_a=J(\ol{f_{\widetilde{\bfU}_{\bullet}^a,\widetilde{\bfU}_{\ast-\bullet}^a}})$
and similarly for $\tilde{J}_b$.  In particular,  to order $O\left(\max\left\{p^{-1/2},p^{-2/(n-2)}\right\}\right)$, if there is a
$q$-sparse covariance under each treatment and there are common
persistent correlations among the $q$ variables the
$$E[N^{\ab}]=\kappa_n^{\ab}\left(1+O\left(\frac{q(q-1)}{p(p-1)}\right) \right),$$
while if only one of the treatments, say treatment $a$, produces
$q$-sparse covariance
$$E[N^{\ab}]=\kappa_n^{\ab}\tilde{J}_b
\left(1+O\left(\frac{q(q-1)}{p(p-1)}\right) \right).$$ In
particular, in the latter case to order $O\left(\max\left\{(q/p)^2, p^{-1/2},p^{-2/(n-2)}\right\}\right)$ the simple
product representation (\ref{eq:NaNb}) holds.

\section{Numerical experiments}
\label{sec:applications}

To illustrate the practical utility of the theory developed in the
previous sections 
we present two numerical studies.  First simulations were performed
that show our false positive rate approximations give accurate
finite $p$ approximations to empirically determined error rates in a
sparse example.
Second, these approximations are used to perform correlation screening
on experimental gene expression microarray data.

\subsection{Simulation results}

We used the asymptotic theory to specify suitable correlation
thresholds that ensure specified familywise error rates (FWER): false
positives (Type I) and false negatives (Type II).  We simulated a
problem of persistent correlation screening over a pair of treatments
for the presence of a few and strongly correlated variables in a
nearly diagonal covariance matrix.  The two treatments were balanced
$n_a=n_b$, the rows of $\mathbb X$ were i.i.d. multivariate normal
and the covariance matrix was diagonal except for a $2\times 2$ block
corresponding to a pair of correlated variables.

For given $p$ and $n_a$, $n_b$, the approximation
to $P(N^{\ab}>0)$ given in Prop. \ref{prop1persistent} was used to
select thresholds $\rho_p^a, \rho_p^b$ that guarantee a Type I FWER of
level $\alpha$.  Once this threshold was determined, the Type II FWER
was approximated using a bias corrected normal approximation to the
Fisher-Z transformation of the non-zero correlations: $Z_{ij}=\half
\log\frac{1+\r_{ij}}{1-\r_{ij}}$: for $n$ the number of samples
$Z_{ij}$ is approximately normally distributed with mean and variance
\cite{Anderson:03}
$$ E[Z_{ij}]=\half \log\frac{1+\rho_{ij}}{1-\rho_{ij}} + \rho_{ij}/(2(n-1)), \;\; \var(Z_{ij}) =(n-3)^{-1}$$

These  approximations to Type I and Type II error
rates were combined to produce Table \ref{table3}. This table
illustrates how one can use the theory to predict the required sample
sizes and the required threshold to achieve a desired false
positive rate $\alpha$.  The minimal detectable correlation is defined
using the aforementioned theoretical FWER approximations as the minimum value of the true correlation for which the presence of a persistent correlation is detected with probability at least $\beta$
and false alarm probability $\alpha$. For example, with the $p=500$
variables assumed in generating the table, at least $n=35$ samples are
required for reliable detection of a persistent magnitude correlation
less than or equal to $\rho_1=0.77$ at the prescribed $(\alpha, \beta)
= (0.01,0.8)$ false positive and true positive levels.

\begin{table}
\begin{center}
\begin{tabular}{|c|c|c|c|c|c|} \hline
$n\diagdown \alpha $ & 0.010 & 0.025 & 0.050 & 0.075 & 0.100\\ \hline
 10 &$  0.98\backslash   0.96$ & $  0.98\backslash   0.96$ & $  0.98\backslash   0.95$ & $  0.98\backslash   0.95$ & $  0.98\backslash  0.95$ \\ \hline
 15 &$  0.94\backslash   0.89$ & $  0.94\backslash   0.88$ & $  0.93\backslash   0.87$ & $  0.93\backslash   0.87$ & $  0.93\backslash  0.87$ \\ \hline
 20 &$  0.89\backslash   0.82$ & $  0.89\backslash   0.81$ & $  0.88\backslash   0.80$ & $  0.88\backslash   0.80$ & $  0.88\backslash  0.79$ \\ \hline
 25 &$  0.85\backslash   0.76$ & $  0.84\backslash   0.75$ & $  0.84\backslash   0.74$ & $  0.83\backslash   0.74$ & $  0.83\backslash  0.73$ \\ \hline
 30 &$  0.81\backslash   0.72$ & $  0.80\backslash   0.70$ & $  0.79\backslash   0.70$ & $  0.79\backslash   0.69$ & $  0.79\backslash  0.69$ \\ \hline
 35 &$  0.77\backslash   0.67$ & $  0.76\backslash   0.66$ & $  0.76\backslash   0.65$ & $  0.75\backslash   0.65$ & $  0.75\backslash  0.64$ \\ \hline
\end{tabular}
\end{center}
\caption{Minimum detectable correlation and level-$\alpha$ threshold
  (given as entry $\rho_1 \backslash \, \rho$ in table) for persistent
  correlation screening as a function of number of samples $n$ (rows)
  and familywise false positive level $\alpha$ (columns) for $p=500$
  and $\beta=0.8$. The number of samples in each treatment is
  identical ( $n=n_a=n_b$). The false positive rate approximation in
  Prop. \ref{prop1persistent} was used to determine the required
  level-$\alpha$ threshold $\rho$. With this value of $\rho$ the
  minimum detectable correlation $\rho_1$ was determined using a bias
  corrected normal approximation to the Fisher-Z transformation of the
  sample correlation.}
\label{table3}
\end{table}

\begin{figure}
\begin{center}
\includegraphics[height=7.8cm]{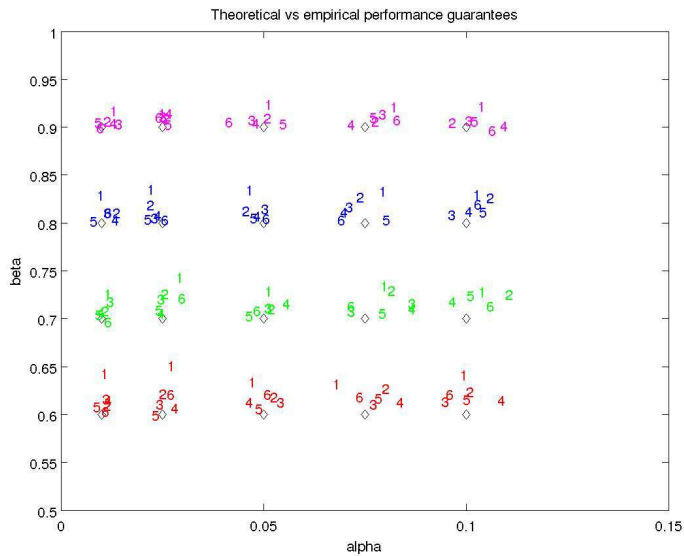}
\end{center}
\caption{Comparison between predicted (diamonds) and actual (integers)
  operating points $(\alpha, \beta)$ for persistent correlation screening
  thresholds determined by the theory used to generate Table \ref{table3}.
  Each integer is located near an operating point and indexes the
  sample size $n$ over the six values $n=10,15,20,25,30,35$. These
  numbers are color coded according to the target value of $\beta$. }
\label{fig:grid3}
\end{figure}

Next we assess the fidelity of the familywise error predictions in
Table \ref{table3} by comparing them to empirical error rates
determined by simulation. To obtain the empirical values a set of
tables like Table \ref{table3} was generated for each targeted value
of $\beta$, ranging from $0.6$ to $0.9$, and the obtained predicted
threshold value $\rho$ was used to screen the sample correlation
matrix.  We simulated 4000 replicates to construct relative
frequencies of empirical false positive rates $\hat{\alpha}$ and
empirical true positive ($\hat{\beta}$) rates for the same parameters
$p,n$ as were used to generate the analytical predictions in the
tables.  Figure \ref{fig:grid3} shows the predicted ($\alpha,\beta$)
operating points (diamonds) and actual ($\alpha,\beta$) operating
points (integers), determined by simulation for different values of
$n$.  Figure \ref{fig:grid3} demonstrates that our asymptotic
predictions are accurate for relatively large values of $\alpha$,
small values of $n$, and finite $p$.
\subsection{Experimental results}

We applied the correlation screening theory to a dataset downloaded
from the public Gene Expression Omnibus (GEO) NCBI web site
\cite{GEOBaty2006}. This data was collected and analyzed by the
authors of \cite{baty2006analysis}.  The dataset consists of 108
Affymetrix HU133 Genechips containing $p=22,283$ gene probes
hybridized from peripheral blood samples taken from 6 individuals at 5
time points (0,1,2, 4 and 12 hours) on four independent days under
$m=4$ treatments: intake of alcohol, grape juice, water, or red wine.
According to the GEO Summary of the author's analysis
of this data: ``Results may contribute to elucidating the mechanisms
underlying the cardioprotective effects of red wine.''

\begin{figure}
\begin{center}
\includegraphics[height=4cm]{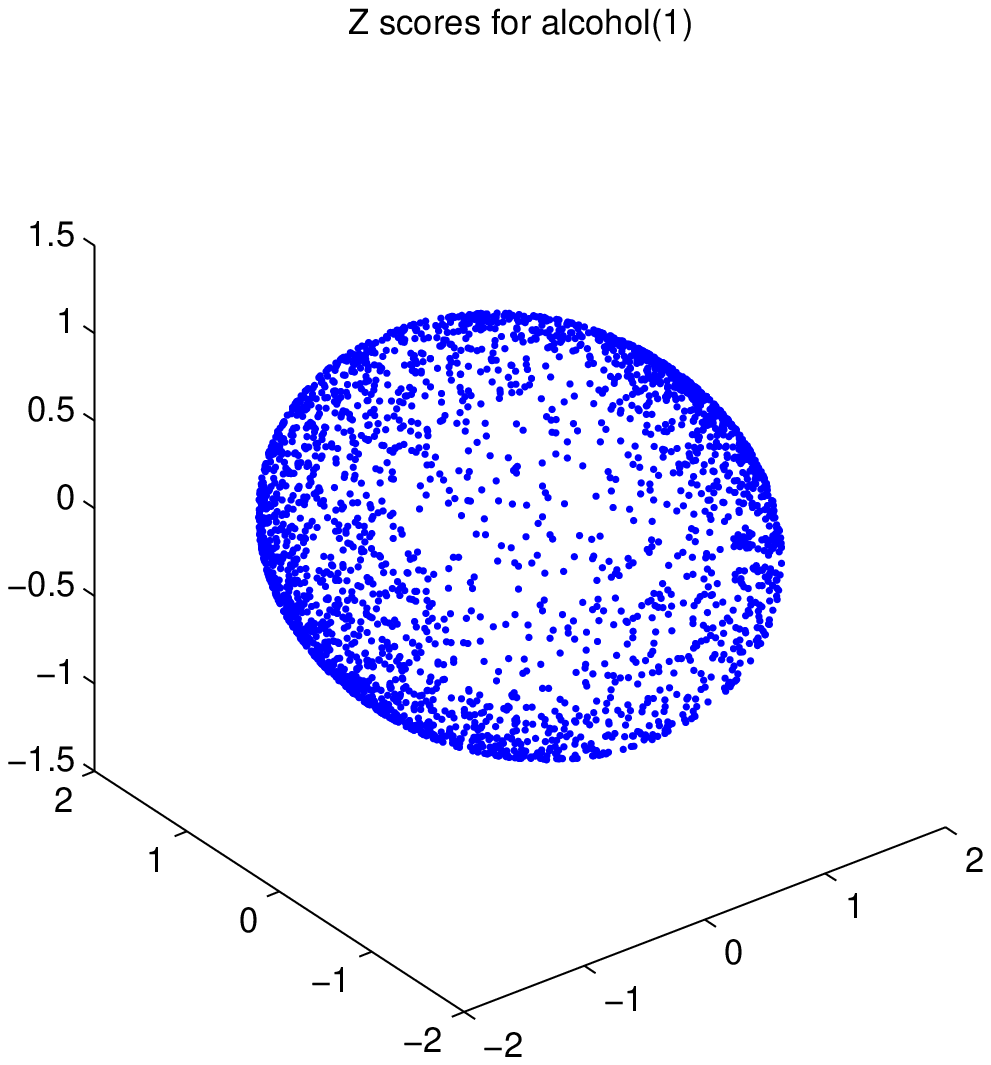}\;\;\;
\includegraphics[height=4cm]{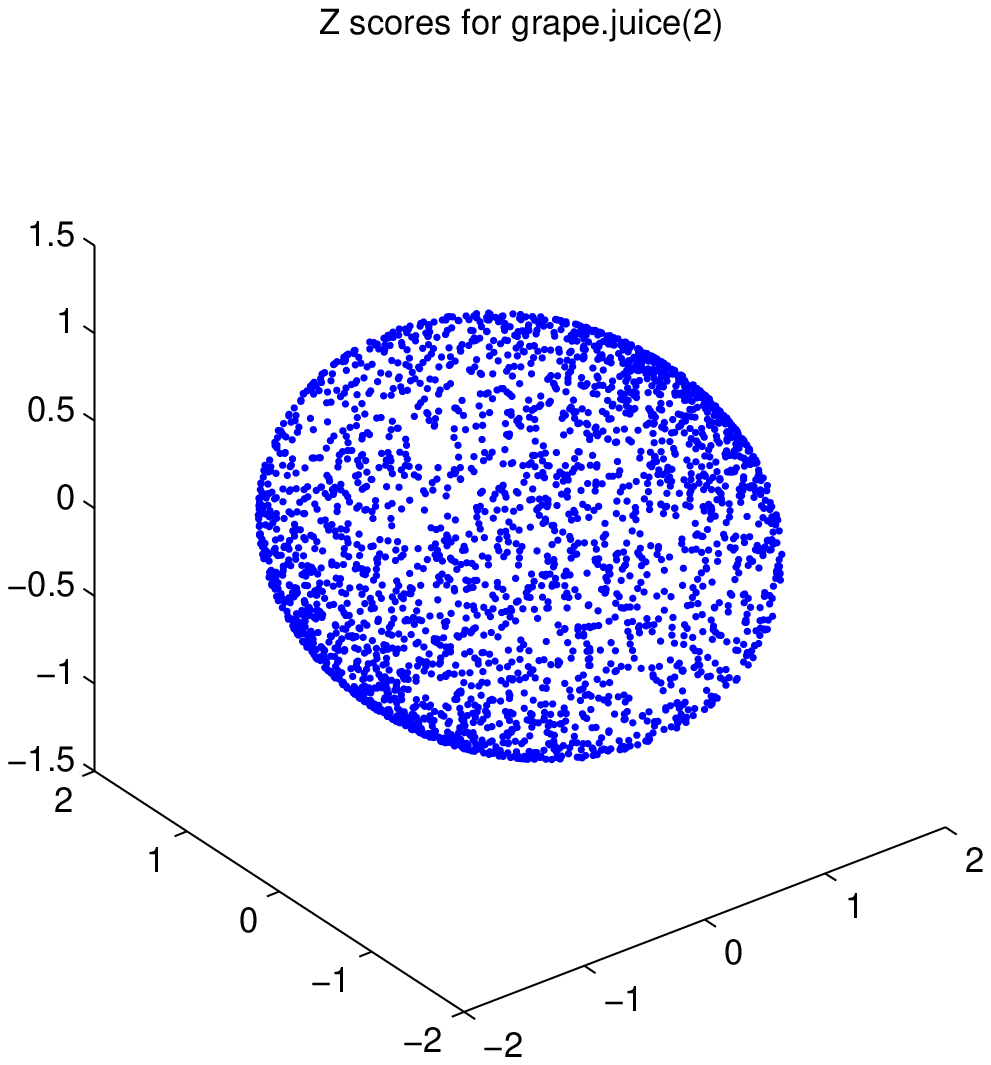}\;\;\; \\
\includegraphics[height=4cm]{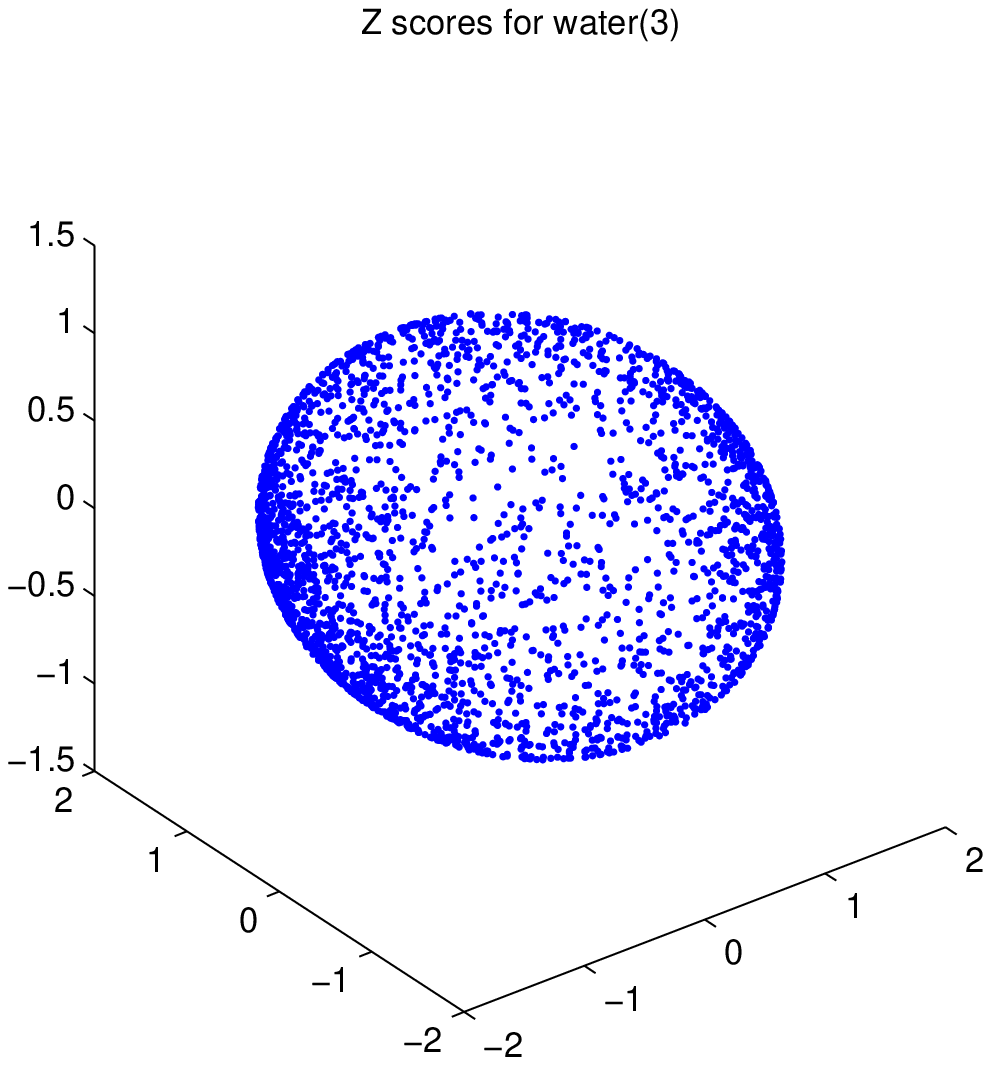}\;\;\;
\includegraphics[height=4cm]{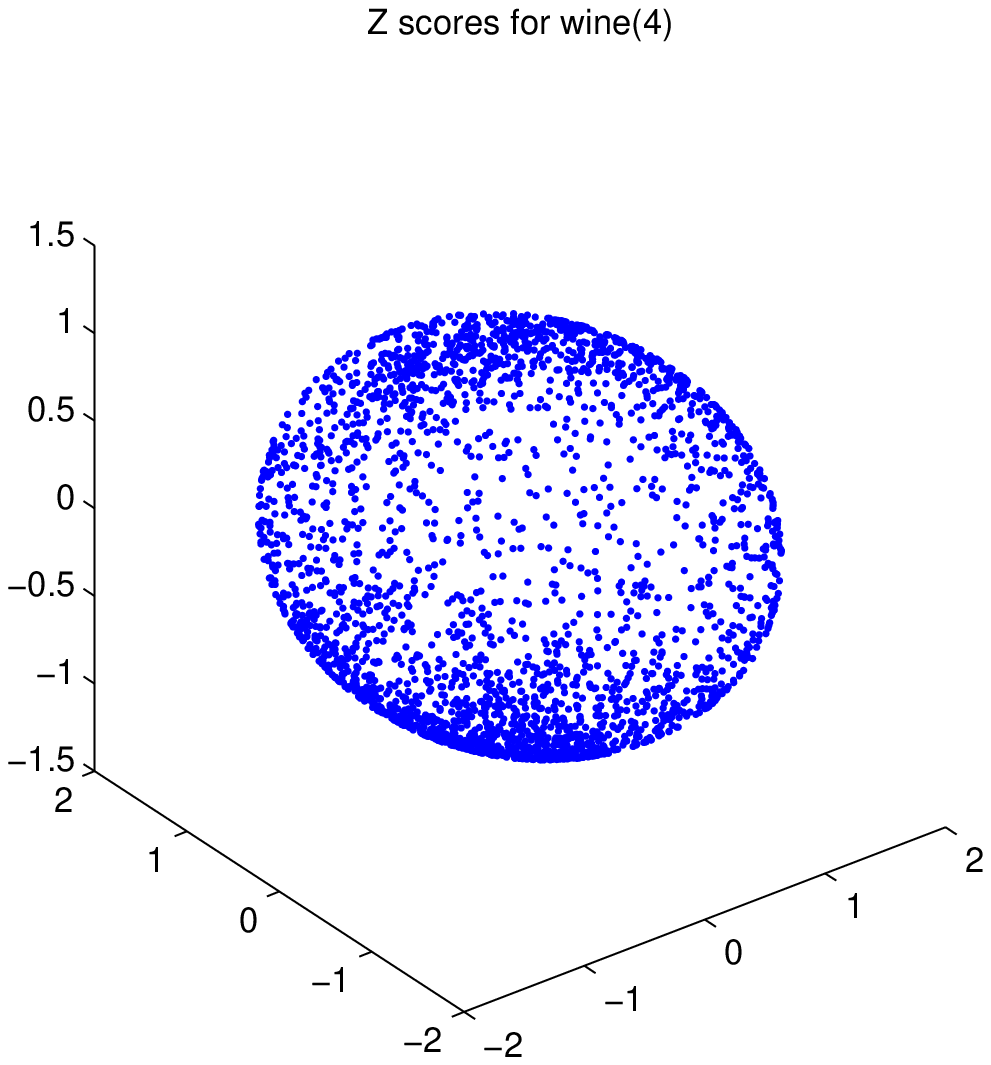}\;\;\;
\end{center}
\caption{3-dimensional projections of the U-scores for the
  experimental beverage data under each of the treatments 1,2,3,4. For
  visualization the 22,238 variables (gene probes) were downsampled by
  a factor of 8 and a randomly selected set of four samples in each
  treatment were used to produce these figures. 
}
\label{fig:quad}
\end{figure}

After removing samples taken at pretreatment baseline (time 0) there remained
$n=87$ samples distributed over the treatments as:
$n_1=20$ (alcohol), $n_2=22$ (grape juice), $n_3=23$ (water), and
$n_4=22$ (wine).  Figure \ref{fig:quad} gives a visualization of the
U-scores for each treatment. Observe that the U-scores display
non-uniformity on the sphere $S_2$. We applied correlation screening
to the data as follows.  As the numbers of samples differ in each
treatment we constrained the screening thresholds to equalize the four
per-treatment auto-screening error rates, as explained in
Sec. \ref{sec:correlationscreening}.

\begin{table}
\begin{center}
\begin{tabular}{|c||c|c|c|c|c|c|}  \hline
$\{1\},\{2\},\{3\},\{4\}$ & & 51 & 52 & 96 & 518 &  \\  \hline
$\{1,2\},\{1,3\},\{1,4\},\{2,3\},\{2,4\},\{3,4\}$ &493 & 748 & 1069 & 677 & 864 & 1445  \\  \hline
$\{2,3,4\},\{1,3,4\},\{1,2,4\},\{1,2,3\}$ & & 2242 & 2530 & 1893 & 1690 &  \\  \hline
$\{1,2,3,4\}$ & & & 3313 &  &  &  \\  \hline
\end{tabular}
\end{center}
\caption{Number of genes discovered by auto-screening (top row) and
  persistency screening (lower three rows) for various combinations of
  treatments in the experimental data. Auto-screening threshold
  determined using our approximation to Type I error of level
  $10^{-5}$. }
\label{table2}
\end{table}

There are $2^4-1$ possible auto-screening and persistency-screening
analysis combinations that can be performed over the 4 treatments
$\{1,2,3,4\}$. Using our approximation to false positive rate we
fixed Type I FWER at level $10^{-5}$ and determined the 4
auto-screening thresholds and the 11 sets of persistency screening
thresholds. Correlation screening
was performed on the sample correlation matrix of all 22,238 gene
probes. These thresholds resulted in 15 different sets of discoveries
in relative numbers shown in Table \ref{table2}.

\begin{figure}
\begin{center}
\includegraphics[height=12cm]{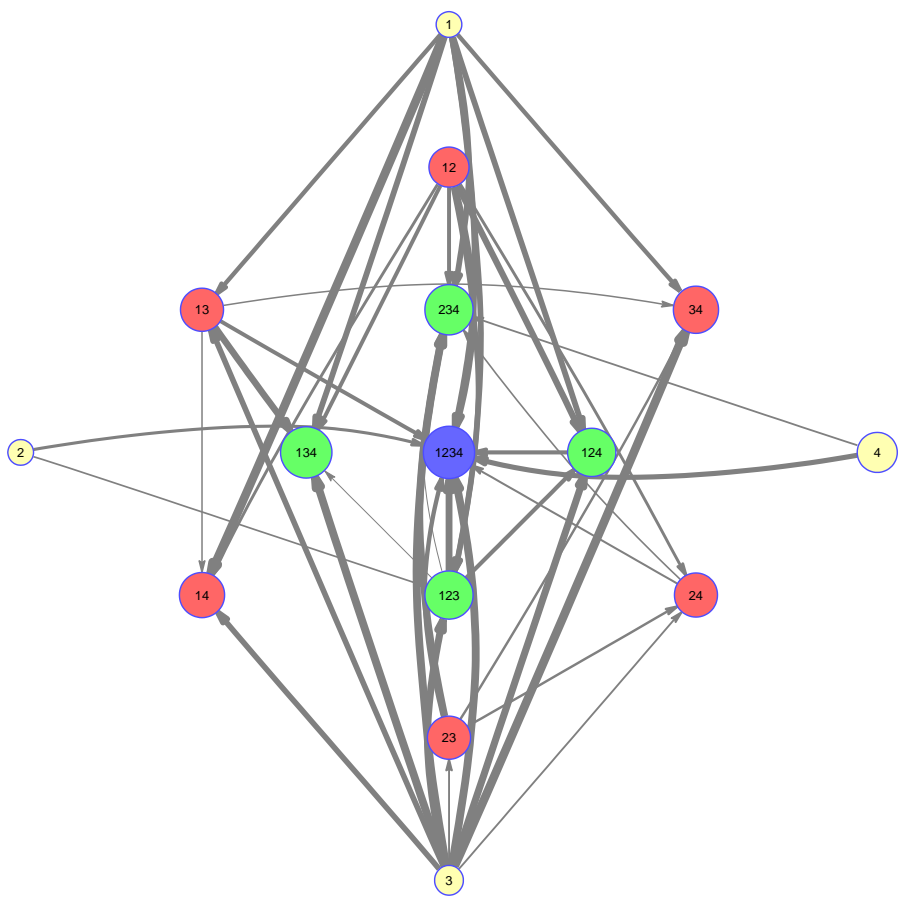}
\end{center}
\caption{Set-inclusion graph between genes discovered by correlation
  screening in various combinations of treatments. Size of node is
  proportional to the log of number of associated correlation
  screening discoveries given in Table \ref{table2}. A directed edge
  from node $i$ to node $j$ exists if at least 90\% of the genes
  discovered in node $i$ are also discovered in node $j$ and the
  thickest edges indicate 100\% set inclusion. The asymmetry of
  diagram indicates that treatments have different effects on gene
  expression. The paucity of edges to and from grape juice (``2'') and
  wine (``4'') indicates that most of the genes discovered in
  auto-screening are not persistent across treatments.}
\label{fig:trtmntgraph}
\end{figure}

To explore the relations between the different sets of genes
discovered in each screen we plot a directed set-inclusion graph in
Fig. \ref{fig:trtmntgraph}. The sizes of the 15 nodes correspond to
the length of the list of discovered genes at FWER $10^{-5}$ under the
persistency screening combination that is indicated by the node
label. The nodes are arranged in 3 concentric rings with an inner ring
corresponding to higher degree of persistency (persistency over more
treatments) than an outer ring. Edges are shown only between nodes for
which at least 90\% of the genes in one node is a subset of the other
node and thickest edges correspond to 100\% set inclusion. There are
no edges between different auto-correlation screens (nodes labeled
1,2,3,4). Note also the preponderance of directed edges with arrows
pointing from outer rings towards inner rings as as contrasted with
edges between nodes on the same ring or pointing to outer rings. As
compared to the other three treatments, treatment 2 (water) generates
a lower proportion of auto-correlation screening genes that are also
persistent genes.


\begin{figure}
\begin{center}
\includegraphics[height=12cm]{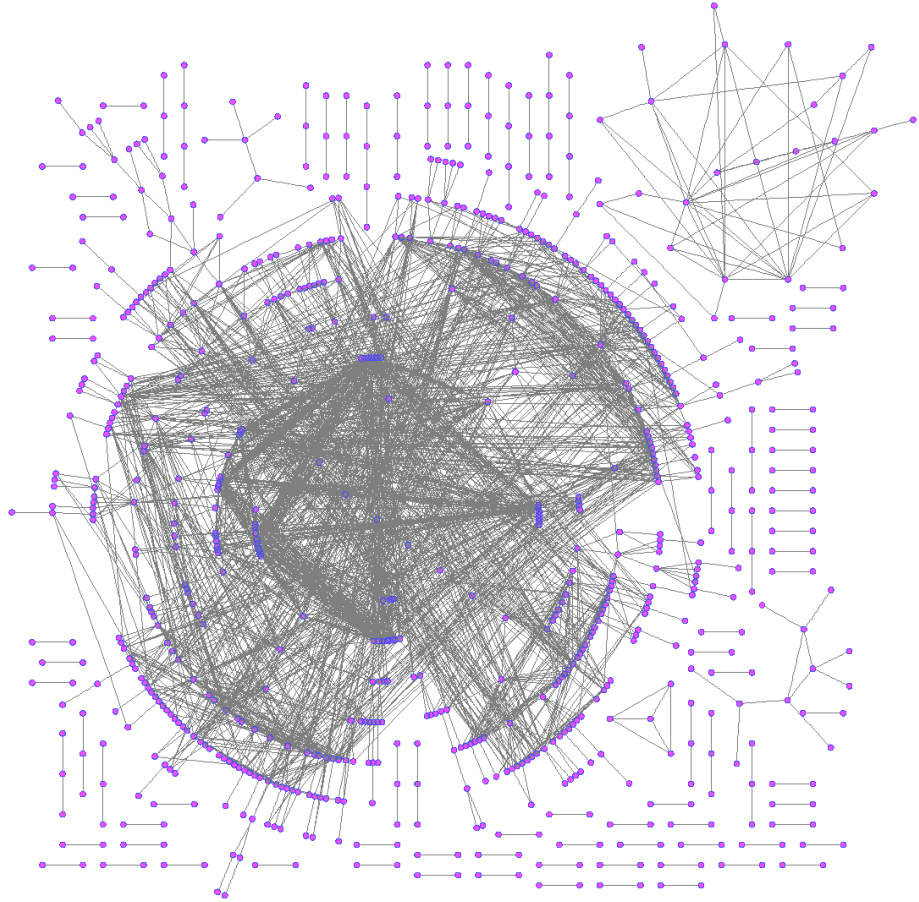}
\end{center}
\caption{774 gene subnetwork of the 3313 gene persistent-correlation
  network across all four treatments corresponding to the last row of
  Table \ref{table2}. Two nodes in this network are linked by an edge
  if for all 4 treatments their sample correlation is above the
  $10^{-5}$ FWER correlation-screening threshold.}
\label{fig:all4}
\end{figure}

In Figure \ref{fig:all4} we show a 774 node subnetwork of the
correlation network corresponding to the 3313 discoveries of genes
whose correlation persists over all four treatments. Two genes in this
subnetwork are connected by an edge only if the sample correlation
between them persists over all four treatments. Thus, as contrasted to
the original 3313 node network of genes having any correlation that
persists over treatments (persistent nodes), Fig. \ref{fig:all4} shows
the subnetwork of genes whose mutual correlations persist (persistent
edges). Observe the presence of a giant component of 516 genes shown
in the figure as the central connected component.


%

\section{Conclusions}

We have presented theory that yields asymptotic approximations
for large scale correlation screening within a single treatment and
across multiple treatments. We obtained expressions for the mean
number of discoveries that depend on Bhattacharyya divergences \cite{Basseville:SP89}.
Expressions for phase transition thresholds were established.  The
theory applies to large scale screening of sample correlation when the
true correlation is sparse or approximately sparse. Put another way,
the theory applies to screening for star motifs in a sparse graph
associated with a thresholded sample correlation matrix. This theory
can be extended to screening more general correlation motifs,
e.g. triangles, chains, and higher order transitive correlations. It
can also be extended to screening sparse partial correlation matrices.

\section*{Supplemental Materials}
\begin{description}
\item[Proofs of propositions, lemmas and corollary]	Proofs of Proposition~\ref{prop1}, \ref{prop1cross} and \ref{prop1persistent}; definitions for proofs, a fundamental lemma, and a corollary; 
\end{description}






\appendix
\setcounter{page}{1}
\pagestyle{fancy}
\fancyhf{} 
\fancyfoot[C]{S-\thepage}
\renewcommand{\headrulewidth}{0pt}

\renewcommand{\thefigure}{\thesection.\arabic{figure}}
\setcounter{figure}{0}

\renewcommand{\thetable}{\thesection.\arabic{table}}
\setcounter{table}{0}

\section*{\huge{Supplemental Materials}}

\section{Proofs of Propositions}

\subsection{Definitions and fundamental lemma}

Here we give the principal definitions used in this Appendix/Supplemental Section.

\noindent{\bf Definitions:}
In the paper we defined averaged densities of one or two variables such as $\ol{f_{\bfU_\ast}}$, $\ol{f_{\bfU^a_\ast,\bfU^b_\bullet}}$ and $\ol{f_{\bfU^a_\bullet,\bfU^b_{\ast-\bullet}}}$. For averages over more than two indices, required in the proofs developed below, we introduce the following notation for $k$-fold averaging. For fixed integer $i$ define
\begin{equation}
\begin{split}
\avg_{i_1 \neq \cdots \neq i_k}
f_{\bfU_{i_1},\ldots,\bfU_{i_k},\bfU_{i}} (\bfu_1, \ldots, \bfu_k,\bfv) &=
(p(p-1)\cdots (p-k+1))^{-1} \\
& \quad \times \sum_{\nntstile{i_1 \neq \cdots \neq i_k\neq i}{i_1,\ldots,i_k}}
f_{\bfU_{i_1},\ldots,\bfU_{i_k},\bfU_{i}} (\bfu_1, \ldots, \bfu_k,\bfv)
, \label{eq:avgdef}
\end{split}
\end{equation}
\noindent and similarly for $\avg_{i_1 \neq \cdots \neq i_k}
f_{\bfU_{i_1},\ldots,\bfU_{i_k}|\bfU_{i}}$.  When all of the variables
$\bfU_{i_1}$ are from the same treatment, the indices $i_1, \ldots,
i_k$ run over the range $1,\ldots,p$ and exclude the index $i$. When
there are two treatments, as in $\avg_{i_1 \neq \cdots \neq i_k}
f_{\bfU_{i_1}^b,\ldots,\bfU_{i_k}^b|\bfU_{i}^a}$, the indices $i_1,
\ldots, i_k$ run over the same range but include $i$.

Thus we have, for example, $\avg_{i,j}\{{f_{\bfU^a_i,\bfU^b_j}\}}=\ol{f_{\bfU^a_\ast,\bfU^b_\bullet}}$ and
$\avg_{i\neq j}\{{f_{\bfU_i,\bfU_j}\}}=\ol{f_{\bfU_{\bullet},\bfU_{\ast-\bullet}}}$.
When there is no risk of confusion,  we will   write the averaging
operator $\avg_{i_1 , \ldots ,i_k}$ instead of $\avg_{i_1 \neq \cdots \neq i_k}$.

Define the least upper bound $M_{k|1}$ on any $k$-th order conditional U-score density\\ 
$f_{\bfU_{i_1},\ldots, \bfU_{i_k}|\bfU_{k+1}}(\bfu_{1},\ldots,\bfu_{k}|\bfu_{k+1})$
\be
M_{k|1} &=&\max_{i_1 \neq \cdots \neq i_{k+1}}\left\| f_{\bfU_{i_1},\ldots, \bfU_{i_k}|\bfU_{k+1}}\right\|_{\infty}
\label{eq:Mk1def},
\ee
where for any function $g(\bfu_1,\ldots,\bfu_k)$ of $\bfu_i \in S_{n-2}$, $i=1, \ldots, k$,  $\|g\|_{\infty}$ denotes the sup norm
$$\|g\|_{\infty}=\sup_{\bfu_{1},\ldots, \bfu_{k} \in S_{n-2}\times
  \cdots \times S_{n-2}}|g(\bfu_1,\ldots,\bfu_k)|.$$ Similarly
define $M_{k|2}$ as:
\be
M_{k|2} =\max_{i_1 \neq \ldots \neq i_{k+2}}\left\|
f_{\bfU_{i_1},\ldots, \bfU_{i_k}|\bfU_{i_{k+1}}, \bfU_{i_{k+2}}}\right\|_{\infty}. \label{eq:Mk2def}
\ee

Define the maximal gradient of the average
pairwise density
%
%
\be \dot{M}_{1|1}=\max_{i\neq j} \sup_{\bfv \in 
  S_{n-2}} \left\|\left.  \nabla_{\bfu} f_{\bfU_i|\bfU_j}(\bfu|\bfv) \right|_{\bfu=\bfv} \right\|_2
 \label{eq:mdotdef},
\ee
where $\nabla_{\bfu} = [ \partial/\partial u_1,
  \ldots,\partial/\partial u_{n-1}]^T$ is the gradient operator.  

For two treatments $a,b$ we define the above quantities analogously except that
the single treatment U-score distribution is 
replaced by the two treatment distribution 
$f_{{\mathbb U}^a, {\mathbb U}^b}$. For example $M_{k|1}$ and $\dot{M}_p$ become
\be
M_{k|1}^{ab} &=&\max_{i_1 \neq \cdots \neq i_{k+1}}\left\| f_{\bfU_{i_1}^a,\ldots, \bfU_{i_k}^a|\bfU_{i_{k+1}}^b}\right\|_{\infty}
\label{eq:Mk1abdef},
\ee
and
\be \dot{M}_{1|1}^{ab}= \max_{i\neq j} \sup_{\bfv \in 
  S_{n-2}} \left\|\left.  \nabla_{\bfu} f_{\bfU_i^a|\bfU_j^b}(\bfu|\bfv) \right|_{\bfu=\bfv} \right\|_2.
 \label{eq:mdotabdef}
\ee

\noindent{\bf Weak dependency coefficients}

For a single treatment, let $\delta_i$ denote the degree of node $X_i$
in the population correlation graph over $\bX=[X_1, \ldots, X_p]$. For
given integer $k$, $0\leq k <p$, define \be \calN_k(i)=\argmax_{j_1
  \neq \cdots \neq j_{\min(k,\delta_i)} \neq i} \sum_{l=1}^{\min(k,\delta_i)}
|\rho_{ij_l}|
\label{eq:Nki}
. \ee When $k\geq \delta_i$ these are indices of the nearest neighbors of
$X_i$ amongst $\{X_j\}_{j\neq i}$. When $k<\delta_i$ these are the
$k$-nearest neighbors ($k$-NN) of $X_i$.  For a pair of U-scores
$\bU_i,\bU_j$ define the $p-2-k$
``complementary $k$ nearest neighbors''
$\bU_{A_k(i,j)}=\{\bU_l: l \in A_k(i,j)\}$ where \be A_{k}(i,j) =
\left(\calN_k(i) \cup \calN_k(j)\right)^c -\{i,j\}
\label{eq:Akij},
\ee
with $A^c$ denoting set complement of $A$. The complementary
$k$-NN's from $\bU_i,\bU_j$ include all scores
outside of their respective  $k$-nearest-neighbor regions.
For $i\neq j$ the dependency coefficient between $\bU_i,\bU_j$ and
their complementary $k$-NN's is defined as
\be
\Delta_{p,k}(i,j)= \left \|(f_{\bU_i,\bU_j|\bU_{A_k(i,j)}}-f_{\bU_i,\bU_j})/f_{\bU_i,\bU_j}
\right\|_{\infty}.
\label{eq:deltapij}
\ee

For two treatments $a,b$ let $\delta_i^a,\delta_i^b$ be the degrees of vertices $X_i^a,X_i^b$, respectively, in the population cross-correlation graph having an edge between $X_i^a$ and $X_j^b$ when $\rho_{ij}^{ab}\neq 0$. Similarly to (\ref{eq:Nki}) define the indices of the $k$-nearest neighbors  of $X_i^a$ among $\{X_j^b\}_j$:
\be
\calN_k^{b|a}(i)=\argmax_{j_1 \neq \cdots
  \neq j_{\min(k,\delta_i^a) }} \sum_{l=1}^{\min(k,\delta_i^a)} |\rho_{ij_l}^{ab}|
  \label{eq:Nkabi}
\ee
and similarly define $\calN_k^{a|b}(i)$ by replacing $\delta_i^a$ with $\delta_i^b$ and $\rho_{ij_l}^{ab}$ with   $\rho_{ij_l}^{ba}$. In analogy to (\ref{eq:Akij}) define
\be
A_{k}^{ab}(i,j) &=& \left\{(l,m): l\in \left(\calN_k^{a|b}(j)\right)^c -\{i\}, m \in \left(\calN_k^{b|a}(i)\right)^c-\{j\}\right\}
\label{eq:Akabij}.
\ee For a pair of U-scores $\bU_i^a,\bU_j^b$ the complementary
$k$-NN's in treatments $a$ and $b$ are $\{\bU_l^a,\bU_m^b\}_{(l,m)\in
  A_{k}^{ab}(i,j)}$.  The cross-dependency coefficient between
$\bU_i^a,\bU_j^b$ and the complementary $k$-NN's is defined as \be
\Delta_{p,k}^{ab}(i,j)= \left \|
(f_{\bU_i^a,\bU_j^b|\{\bU_l^a,\bU_m^b\}_{(l,m)\in
    A_{k}^{ab}(i,j)}}-f_{\bU_i^a,\bU_j^b})/f_{\bU_i^a,\bU_j^b}\right\|_{\infty}
\label{eq:deltapabij} .
\ee

Finally, let the average U-score weak dependency and weak cross-dependency coefficients be given by arithmetic averages
\be
\|\Delta_{p,k}\|_1= (p(p-1)/2)^{-1}\sum_{i< j} \Delta_{p,k}(i,j) , \;\;\; \|\Delta_{p,k}^{ab}\|_1= p^{-2}\sum_{i, j=1}^p \Delta_{p,k}^{ab}(i,j).
\label{eq:Deltapdefavg}
\ee

The average weak dependency coefficients (\ref{eq:Deltapdefavg}) are a
natural measure of sparsity and weak dependence. For example, assume
that there is no vertex of degree greater than $k$ in the population
correlation graph associated with $\bX$, that $\bX$ has an
elliptical distribution and that the rows of the data matrix
$\mathbb X$ are i.i.d. Then $\|\Delta_{p,k}\|_1=0$. Similarly, if the rows of
$[\mathbb X^a,\mathbb X^b]$ are i.i.d. elliptically distributed and no node
in the population cross-correlation graph has vertex degree exceeding
$k$ then $\|\Delta_{p,k}^{ab}\|_1=0$.

\subsection{Proofs of Propositions}

The proofs of Props. \ref{prop1}-\ref{prop1persistent} will use several
fundamental results gathered in the following lemma.

\begin{lemmas}
Let $\mathbb X_p$ be a  $n \times p$ data matrix and
let  $\{\bfU_i\}_{i=1}^{p}$ be the U-scores extracted from the columns of $\mathbb X_p$.
Assume that the joint U-score density is bounded.  Define $\phi_{ij}$ the indicator function of the event $|r_{ij}| \geq \rho$ where $r_{ij}=\bfU_i^T\bfU_j$ is the sample-correlation coefficient and $0\leq\rho\leq 1$.
Then for any $i_1, \ldots, i_k \in \{1, \ldots , p\}$,
$i_1 \neq \cdots \neq i_k \neq i$, $k\in \{1, \ldots, p-1\}$,
\be E\left[\prod_{j=1}^k \phi_{i i_j} \right] &=&\int_{S_{n-2}} d\bfv \int_{A(r,\bfv)} d\bfu_{1} \cdots
\int_{A(r,\bfv)} d\bfu_{k}\;
f_{\bfU_{i_1},\ldots, \bfU_{i_k},\bfU_i}(\bfu_1,
\cdots,\bfu_k,\bfv)
\label{eq:thetamoms}\\
&\leq& P_0^k a_n^k M_{k|1}
\label{eq:thetamomsineq}
\ee
with $P_0=P_0(\rho,n)$ defined in (\ref{eq:Podef}), $a_n=|S_{n-2}|$, and $M_{k|1}$ defined in (\ref{eq:Mk1def}).
In (\ref{eq:thetamoms}) $A(r,\bfv)=C(r,\bfv)\cup C(r,-\bfv)$ is the union of
spherical cap regions on $S_{n-2}$ centered at $\bfv$ and $-\bfv$ with radius
$r=\sqrt{2(1-\rho)}$.

Furthermore, defining $\theta_{i} = (p-1)^{-1} \sum_{\nntstile{j\neq i}{j=1}}^{p} \phi_{i j}$:
\be
|E[\theta_{i}] - P_0 J\left(\ol{f_{\bfU_{i},\bfU_{\ast-i}}}\right)| &\leq& 2  a_n P_0 r \dot{M}_{1|1},
\label{eq:thetafirstmomi}
\ee

When $(p-1)P_0\leq 1 $ we have the following inequality
\be
\left|E[\phi_i]-(p-1)E[\theta_i]\right|\leq
\gamma_{p} \left((p-1)P_0\right)^2.
\label{eq:phithetabnd}
\ee
and, for $i\neq j$,
\be
|E[\phi_{ij}] - P_0 J\left(f_{\bfU_{i},\bfU_{j}}\right)| &\leq&
2  a_n P_0 r \dot{M}_{1|1},
\label{eq:phi_ijdef}
\ee
\be
E[\phi_{ij}]  &\leq& a_n P_0 M_{1|1},
\label{eq:phi_ijbnd}
\ee
and for $i \neq j \neq k\neq l$
\be
E[\phi_{ij}\phi_{jk}]\leq a_n^2 P_0^2 M_{2|1},\;\;\; E[\phi_{ij}\phi_{kl}]\leq a_n^2 P_0^2 M_{2|2}
\label{eq:phi_ijbnd2} .
\ee
\label{lemma:fundamental}
\end{lemmas}

{\it Proof of Lemma \ref{lemma:fundamental}}

Fix $p$. Without any loss we can assume that the indices have been reindexed so
that $i=p$.  The representation (\ref{eq:thetamoms}) follows directly from the fact that $\phi_{ij}$ is the indicator of $\bfU_{j} \in A(r,\bfU_i)$; the event that the magnitude sample correlation between the $i$-th and $j$-th variable exceeds $\rho$, $j\neq i$.
Application of the mean value theorem to the inner integral in
(\ref{eq:thetamoms}), and noting that $|A(r,\bfv)|=a_nP_0$,
with $a_n=|S_{n-2}|$, yields the inequality (\ref{eq:thetamomsineq}).

We next establish (\ref{eq:thetafirstmomi}) and (\ref{eq:phithetabnd}). 
Using the definition of $\theta_i$
and the integral relation (\ref{eq:thetamoms}) for $E[\phi_{ij}]$
\be
E[\theta_i]&=&|A(r,\bfv)|\int_{S_{n-2}} d\bfv
\; \left(\half\ol{f_{\bfU_i, \bfU_{\ast-i}}}(\bfv,
\bfv)+\half\ol{f_{\bfU_i, \bfU_{\ast-i}}}(-\bfv,
\bfv) \right)+\delta_i
\label{eq:firstmomproof}\ee
where 
$\delta_i$ is a residual
that has magnitude upper bounded by $2r\dot{M}_{1|1}$.
To show relation (\ref{eq:phithetabnd}) start with the representation
$\phi_i=\max_{j\neq i}\phi_{ij}$  or, equivalently, $\phi_i=1-\prod_{j\neq i}(1-\phi_{ij})$. Expansion of the product yields the $p$-term series expression
\be
E[\phi_i]= \sum_{j:j\neq i} E[\phi_{ij}] + \sum_{j_1,j_2: j_1<j_2, j_1,j_2 \neq i} E[\phi_{ij_1\phi_{ij_2}}]+ \ldots + E\left[\prod_{j:j\neq i} \phi_{ij}\right] ,
\label{eq:Ephiexpansion}
\ee
where the indices in the summations and the product are indexing over the ranges $1, \ldots, p$. There are $p-1 \choose k$ summands in the $k$-th term on the right of (\ref{eq:Ephiexpansion}) and, by (\ref{eq:thetamomsineq}),  each of these summands is bounded by $P_0^k a_n^k M_{k|1}$.  Therefore, using the definition of $\theta_i$
\be
|E[\phi_i]-(p-1)E[\theta_i]| &\leq& \sum_{k=2}^{p-1} {p-1 \choose k} (P_0 a_n)^k M_{k|1} \nonumber \\
&\leq & \max_{k< p}\{a_n^k M_{k|1}\} \sum_{k=2}^{p-1} {p-1 \choose k} P_0^k
\label{eq:goodbnd}
\ee
Under the assumption $(p-1)P_0 \leq 1$ the sum on the right hand side is bounded by
$((p-1)P_0)^2(e-3/2)$, which establishes (\ref{eq:phithetabnd}). This latter bound
follows from the elementary inequalities
\be
\sum_{k=2}^{G} {G \choose k}  \left(\frac{t}{G}\right)^k \leq \sum_{k=2}^{G} \frac{t^k}{k!} \leq (e-3/2) t^2 , \;\;\; 0\leq t \leq 1.
\label{eq:elementary1}
\ee

Relations (\ref{eq:phi_ijdef}) and (\ref{eq:phi_ijbnd})  are simply recapitulations of (\ref{eq:thetamomsineq}) and (\ref{eq:phi_ijbnd2}) is established analogously.
This finishes the proof of Lemma
\ref{lemma:fundamental}. \qed

\subsection{Proof of Prop. \ref{prop1}}

We divide the proof into two pieces, the first dealing with the mean
number of discoveries (\ref{eq:ENdef}) and the second with the
Poisson limit. Both parts use the following direct consequence of the
expression (\ref{eq:Porhorelation})
\ben
p(p-1)P_0= p(p-1) (n-2)^{-1}\ra_n(1-\rho_p^2)^{(n-2)/2}(1+O(1-\rho_p^2)) ,
\een
so that,  as $p\rightarrow \infty$,
\be
p(p-1)P_0 \rightarrow  e_n\ra_n/(n-2),
\label{eq:Porhorelationapp}
\ee
where $e_n$ is the constant in the rate of convergence of $\rho_p
\rightarrow 1$ that was assumed in Prop. \ref{prop1}. Furthermore, as
$p(p-1)P_0$ converges, $(p-1)P_0$ converges to zero.

By (\ref{eq:phithetabnd}) of Lemma \ref{lemma:fundamental}, when $(p-1)P_0\leq 1$ the number of discoveries $N=\sum_{i=1}^p\phi_i$  has mean that satisfies
\be
\left|E[N]-(p-1)\sum_{i=1}^p E[\theta_i]\right|\leq
p^{-1}\left(p(p-1)P_0\right)^2 \gamma_p .
\label{eq:phithetabnd1}
\ee
Therefore, $E[N]$ converges to $(p-1)\sum_{i=1}^p E[\theta_i]$ with
rate at least $O(p^{-1})$. 

Next consider the difference
$
(p-1)\sum_{i=1}^p E[\theta_i]-p(p-1)P_0 J(\ol{f_{\bfU_{\ast},\bfU_{\ast-\bullet}}})
$.
As $J(\ol{f_{\bfU_{\ast},\bfU_{\ast-\bullet}}})=\sum_{i=1}^pJ(\ol{f_{\bfU_{i},\bfU_{\ast-i}}})$, averaging over $i$ the relation (\ref{eq:firstmomproof}), used to show (\ref{eq:thetafirstmomi})  of Lemma \ref{lemma:fundamental}, provides the bound
\be
\left|(p-1)\sum_{i=1}^p E[\theta_i]-p(p-1)P_0 J(\ol{f_{\bfU_{\ast},\bfU_{\ast-\bullet}}})\right| \leq r  p(p-1) P_0  (2 a_n \dot{M}_{1|1})  ,
\label{eq:ENdefapp2}
\ee
where $r=\sqrt{2(1-\rho)}$.
Combining (\ref{eq:phithetabnd1}) and (\ref{eq:ENdefapp2}) yields
\be
\left|E[N]-(p(p-1)P_0)J(\ol{f_{\bfU_{\ast},\bfU_{\ast-\bullet}}})\right|\leq
p^{-1}\left(p(p-1)P_0\right)^2\gamma_p  +  r_p(p(p-1) P_0) \eta_p,
\label{eq:prop1proofpart1}
\ee
where $\eta_p=2 a_n \dot{M}_{1|1}$. As  $p(p-1) P_0$ converges to
$e_n\ra_n/(n-2)$ and $r_p$ converges to zero, $E[N]$ converges to the stated limit. When $\dot{M}_{1|1} =O(1)$ and $n>4$ the term involving $r_p$ dominates and the bound is of order $O(\sqrt{(1-\rho_p)})=O(p^{-2/(n-2)})$. This completes the first part of the proof.

We next show the stated limit $P(N>0)\rightarrow
1-\exp(-\Lambda)$. Let $\phi_{ij}$ be the indicator of the event
$|r_{ij}|\geq\rho_p$ as defined in Lemma \ref{lemma:fundamental}. Then
$N_e=\sum_{i>j}\phi_{ij}=\half \sum_{i \neq j}\phi_{ij}$
is the number of edges in the thresholded
empirical correlation graph and $N = \sum_{i=1}^p \max_{j:j\neq i}
\phi_{ij}$ is the number of vertices of positive degree.  Since $N=0$
if and only if $N_e=0$: $P(N>0) = P(N_e>0)$.  Thus the stated limit
will follow from: (1) convergence of the distribution of $N_e$ to a
Poisson law with rate $\Lambda=E[N_e]$; (2) convergence of $\Lambda$
to one half of the right hand side of (\ref{eq:ENdef}). Assertion (2)
follows from (\ref{eq:phi_ijdef}) and the obvious identity
$E[N_e]=\avg_{i>j}E[\phi_{ij}]p(p-1)/2$. It remains to show (1).

Define the sets of index pairs $C=\{(i,j): 1\leq i < j\leq
p\}$ and $B_{i,j} = \{(l,m): l\in \calN_k(i), m \in \calN_k(j)\} \cap C$. Observe that $|B_{ij}| \leq k(k-1)/2$.  Let $N^*$ be a
Poisson random variable with rate $\Lambda=E[N_e]$. With these
definitions the Chen-Stein theorem \cite[Thm. 1]{arratia1990poisson}
provides a bound on the total variation distance between the
distribution of $N_e$ and that of $N^*$:
\be \max_A |P(N_e \in A)-P(N^*\in
A)| \leq b_1+b_2+b_3
\label{eq:chenstein}
\ee
where 
$$b_1 = \sum_{(i,j)\in C}\sum_{(l,m) \in B_{ij}}
E[\phi_{ij}]E[\phi_{lm}]
$$
$$b_2 = \sum_{(i,j)\in C}\sum_{(l,m) \in B_{ij}-\{(i,j)\}}
E[\phi_{ij}\phi_{lm}] ,$$
and, for $p_{ij}=E[\phi_{ij}]$,
$$ b_3 = \sum_{(i,j)\in C} E\left[E[\phi_{ij}-p_{ij}|\{\phi_{lm}:(l,m)
    \not \in B_{ij}\cup \{i,j)\}\}]\right].$$
Applying the bound
(\ref{eq:phi_ijbnd}) to the summand of $b_1$ we obtain
$$b_1\leq
\frac{p(p-1)}{2} \frac{k(k-1)}{2} \max_{i<j} E^2[\phi_{ij}]\leq
O(p^2k^2 P_0^2)=O\left((k/p)^2\right),$$
as $p(p-1)P_0=O(1)$.
Likewise, the bound (\ref{eq:phi_ijbnd2}) applied to $b_2$ gives
$$b_2\leq\frac{p(p-1)}{2}  \frac{k(k-1)}{2} \max_{(i,j)\neq (l,m)} E[\phi_{ij}\phi_{lm}]
\leq p^2k^2P_0^2 M a_n^2=O\left((k/p)^2\right),$$
where $M= \max\{ M_{2|1}, M_{2|2}\}$.

Furthermore, with $A_k(i,j)$ the index set defined in (\ref{eq:Akij}),
\ben
&&E\left[E[\phi_{ij}-p_{ij}|\{\phi_{lm}:(l,m) \not \in B_{ij}\cup \{(i,j)\}\}]\right]=E\left[E[\phi_{ij}-p_{ij}|\bU_{A_k(i,j)}]\right]\\
&& \hspace{0.4in} =\int_{S_{n-2}^{|A_k(i,j)|}} d\bu_{A_k(i,j)} \int_{S_{n-2}} d\bu_i \int_{A(r,\bu_i)} d\bu_j \\
&&\hspace{0.5in}\left(\frac{f_{\bU_i,\bU_j|\bU_{A_k(i,j)}}(\bu_i,\bu_j|\bu_{A_k(i,j)})- f_{\bU_i,\bU_j}(\bu_i,\bu_j)}{f_{\bU_i,\bU_j}(\bu_i,\bu_j)} \right)
f_{\bU_i,\bU_j}(\bu_i,\bu_j)f_{\bU_{A_k(i,j)}}(\bu_{A_k(i,j)})
\\
&&\hspace{0.4in} \leq a_n P_0 \Delta_{p,k}(i,j) .\een
Hence,   as $b_1+b_2+b_3=O\left(\max\{(k/p)^2,\|\Delta_{p,k}\|_1\right)$
and $k=o(p)$,
(\ref{eq:chenstein}) establishes that $N_e$ converges in distribution to a Poisson random variable. \qed

The rate of convergence of $E[N]$ to $(p-1) \sum_{i=1}^p E[\theta_i]$
specified by (\ref{eq:phithetabnd}) is $O(p^{-1})$, while, when $n>4$,
its rate of convergence to $p(p-1)P_0
J(\ol{f_{\bfU_{\ast},\bfU_{\ast-\bullet}}})$ is dominated by the
slower rate $O(p^{-2/(n-2)})$.  In the case that the rows of $\mathbb X$ are
i.i.d. elliptically distributed with row-sparse covariance matrix $\mathbf
\Sigma$ of degree $k$, the rate of
convergence of the probability $P(N>0)$ to $1-\exp(-\Lambda)$ is at
worst $O\left(\max\{(k/p)^2\}\right)$.

\subsection{Proof of Prop. \ref{prop1cross}}

The technical details for the proof of Prop. \ref{prop1cross} are
similar to those of the proof of Prop. \ref{prop1}.
The main difference is that a discovery ($N^{ab}>0$) occurs when a U-score
$\bfU_j^b$ from treatment $b$ is in the $r$ neighborhood $A(r,
\bfU_i^a)$ of U-score $\bfU_i^a$ from treatment $a$. Therefore, as
contrasted to the auto-screening case, there are $p$ possible
$b$-treatment U-scores that can fall into the neighborhood of
$\bfU_i^a$ instead of the $p-1$ the remaining $a$-treatment U-scores
considered in auto-correlation screening.  Due to this difference, the
factor $p-1$ is replaced by $p$ in all bounds and representations and
the indexing is no longer restricted to distinct indices in $\{\bfU_i^a\}_i$
and $\{\bfU_j^b\}_i$.

The stated limiting expression for $P(N^{ab}>0)$ is established by applying the
Chen-Stein theorem \cite[Thm. 1]{arratia1990poisson} to the number
of edges $N_e=\sum_{i\neq j} \phi_{ij}^{ab}$ in the thresholded empirical cross-correlation graph, where
$\phi_{ij}^{ab}$ is the indicator of the event
$|r_{ij}^{ab}|\geq\rho_p$.  It is easily established that $E[N_e^{ab}]=E[N^{ab}]$.
Define the sets $C=\{(i,j): i,j =1,\ldots,
p\}$ and $B_{i,j}^{ab} = \{(l,m): l\in \calN_k^{a|b}(j), m \in
\calN_k^{b|a}(i)\}$, where $\calN_k^{b|a}(i)$ is defined in (\ref{eq:Nkabi}).  Observe that $|B_{ij}| \leq k^2$ and that the scores
$\{\bU_l^a,\bU_m^b\}_{l,m}$ such that $(l,m)\not\in B_{i,j} \cup \{(i,j)\}$ is
the precisely the set $\{\bU_l^a,\bU_m^b\}_{(l,m)\in A_{k}^{ab}(i,j)}$
where $A_k^{ab}(i,j)$ is given by
(\ref{eq:Akabij}). In analogous manner to the proof of
Prop. \ref{prop1} the three terms $b_1$, $b_2$ and $b_3$ in
(\ref{eq:chenstein}) can be bounded by $b_1\leq p^2 k^2 P_0^2 a_n^2
(M_{1|1}^{ab})^2$, $b_2\leq p^2 k^2 P_0^2 a_n^2\max
(M_{2|1}^{ab}),M_{2|2}^{ab})$ and $b_3 \leq p^2 P_0
a_n\|\Delta_{p,k}^{ab}\|_1 $ where $\|\Delta_{p,k}^{ab}\|_1$ is given by
(\ref{eq:deltapabij}).  Therefore $b_1+b_2+b_3\leq
O(\max\{(k/p)^2,\|\Delta_{p,k}^{ab}\|_1)\})$ and we conclude that if $k
=o(p)$ and $\|\Delta_{p,k}^{ab}\|_1$ converges to zero then $N_e^{ab}$
converges to a Poisson random variable.  Furthermore, from
(\ref{eq:phi_ijdef}) it is easily verified that
$E[N^{ab}]=E[N_e^{ab}]$. Thus, as $N^{\ab}=0$ if and only if $N_e^{\ab}=0$,
$P(N^{ab}>0)=P(N_e^{ab}>0)=1-\exp(-\Lambda^{ab})$ with $\Lambda =
E[N^{ab}]$.  \qed

\subsection{Proof of Prop. \ref{prop1persistent}}


To simplify notation we define $P_{0,a}=P_0(\rho_p^a,n_a)$, $P_{0,b}=P_0(\rho_p^b,n_b)$.
Similarly to the proof of Prop. \ref{prop1}, a direct consequence of the expression (\ref{eq:Porhorelation}) is that  for any $\alpha \in [0,1]$: $p^{-1/2}(p-1)P_{0,a}^\alpha P_{0,b}^{1-\alpha}$ is convergent and therefore $(p-1)P_{0,a}^\alpha P_{0,b}^{1-\alpha}$ converges to zero.

As in  Lemma \ref{lemma:fundamental} define $\phi_i^a=\max_{j\neq i}\phi_{ij}^a$
 the indicator function of the event that in treatment $a$ there is some variable $j \neq i$ whose sample correlation with the $i$-th
 variable  exceeds $\rho_p^a$. Similarly define
 $\phi_i^b$. The total number of persistent discoveries is $N^{\ab} =
 \sum_{i=1}^p \phi_i^a\phi_i^b $ and, since the treatments are independent, $E[N^{\ab}] =
 \sum_{i=1}^p E[\phi_i^a]E[\phi_i^b] $.
 Define the independent random variables $\theta_{i}^a = (p-1)^{-1} \sum_{\nntstile{j\neq i}{j=1}}^{p} \phi_{i j}^a$ and $\theta_{i}^b = (p-1)^{-1} \sum_{\nntstile{j\neq i}{j=1}}^{p} \phi_{i j}^b$.

Consider the difference
 \ben
E[\phi_i^a]E[\phi_i^b]-(p-1)^2E[\theta_i^a]E[\theta_i^b]&=&
(E[\phi_i^a]-(p-1)E[\theta_i^a])
(E[\phi_i^b]-(p-1)E[\theta_i^b])\\
 &&\hspace{-1.0in}+(p-1)E[\theta_i^a](E[\phi_i^b]-(p-1)E[\theta_i^b])+(p-1)E[\theta_i^b](E[\phi_i^a]-(p-1)E[\theta_i^a]) .
 \een
Sum 
over $i$ and  apply inequalities (\ref{eq:phithetabnd}) and (\ref{eq:thetamomsineq}) of Lemma \ref{lemma:fundamental} to obtain, for
$p$ large enough to make $(p-1)P_{0,a}\leq 1$ and $(p-1)P_{0,b}\leq 1$,
%
\begin{equation}
\begin{split}
\left|E[N^{\ab}]-(p-1)^2\sum_{i=1}^p E[\theta_i^a]E[\theta_i^b]\right| &\leq \gamma_p^a\gamma_p^b \left(p^{1/2}(p-1)P_{0,a}^{1/2}P_{0,b}^{1/2}\right)^2p^{-1}\\
& +\left(\eta_p^a\left(p^{1/2}(p-1)P_{0,a}^{2/3}P_{0,b}^{1/3} \right)^3 +\eta_p^b\left(p^{1/2}(p-1)P_{0,b}^{2/3}P_{0,a}^{1/3} \right)^3 \right)p^{-1/2}
\end{split}
\end{equation}where $\gamma_p^a$, $\gamma_p^b$ are defined as  $\gamma_p$ in Lemma \ref{lemma:fundamental} using $M_{k|1}=M_{k|1}^a$ and $M_{k|1}=M_{k|1}^b$, respectively, and $\eta_p^a=a_nM_{1|1}^a\gamma_p^b$, $\eta_p^b =a_nM_{1|1}^b\gamma_p^a$.
As the right hand side of the above equation
is $O(p^{-1/2})$ this establishes that
$$E[N^{\ab}]\rightarrow \lim_{p \rightarrow \infty}
(p-1)^2\sum_{i=1}^p E[\theta_i^a]E[\theta_i^b].$$ By
(\ref{eq:thetafirstmomi}) of Lemma \ref{lemma:fundamental} this limit
is equal to (\ref{eq:avgfubivpersistent}).

It remains to establish the stated limit of the probability
$P(N^{\ab}>0)$.  Similar to the proof of Prop. \ref{prop1}, let
$\phi_{ij}^a$ and $\phi_{ij}^b$ be indicators of the events
$|r_{ij}^a|\geq \rho_p$, $|r_{ij}^b|\geq \rho_p$, respectively. Then
$\phi_i^{\ab} = \max_{j:j\neq i,l:l\neq i} \phi_{ij}^a \phi_{il}^b$
and $N^{\ab}= \sum_{i=1}^p \phi_{i}^{\ab}$. Let
$N_{d^ad^b}=\sum_{i=1}^p \sum_{j:j \neq i} \sum_{l:l\neq i}
\phi_{ij}^a \phi_{il}^b= \sum_{i=1}^p d_i^ad_i^b$ where $d_i^a$ and
$d_i^b$ denote the degrees of vertex $i$ in the respective thresholded
empirical correlation graphs associated with each treatment. We will
show that $N_{d^ad^b}$ is asymptotically Poisson distributed with
rate $\Lambda=E[N^{\ab}]$. Since $N^{\ab}=0$ if and only if
$N_{d^ad^b}=0$ this will establish the stated limiting expression for
$P(N^{\ab}>0)$.

First we establish that $E[N_{d^ad^b}]$ converges to the same limit as does
$E[N^{\ab}]$.  Since
the treatments are independent
\be E[N_{d^ad^b}]= \sum_{i=1}^p \sum_{j:j \neq i} E[\phi_{ij}^a]
\sum_{l:l\neq i} E[ \phi_{il}^b] .
\label{eq:persistmeanij}
\ee Invoking (\ref{eq:phi_ijdef}) from Lemma \ref{lemma:fundamental},
\ben |E[\phi_{ij}^a]- P_{a,0} J\left(f_{\bfU^a_{i},\bfU^a_{j}}\right)| \leq 2 a_n P_{a,0} r_p^a \dot{M}^a_{1|1}, \een 
where $r_p^a=\sqrt{2(1-\rho_p^a)}$, and
likewise for $E[\phi_{ij}^b]$.  Therefore, from
(\ref{eq:persistmeanij}),
\ben E[N_{d^ad^b}]=p(p-1)^2 P_{a,0}P_{b,0}
\left(p^{-1}\sum_{i=1}^p
J\left(f_{\bfU^a_{i},\bfU^a_{\bullet-i}}\right)J\left(f_{\bfU^b_{i},\bfU^b_{\bullet-i}}\right)
+ O(r_p)\right) \een where 
$O(r_p)\rightarrow 0$ as $\rho_p^a,\rho_p^b\rightarrow
1$.
Therefore $E[N_{d^ad^b}]$  converges to the limit on the right side of
(\ref{eq:avgfubivpersistent}).

Define $C=\{(l,m,n): 1\leq l,m,n \leq p,  m \neq l, n\neq l)\}$.
For given integer $k$, define the index set $$B_{i,j,l}= \{(l,m,n):
l\in \calN_k^{a}(i) \cup \calN_k^{b}(i), m \in \calN_k^{a}(j), n\in
\calN_k^{b}(l)\}\cap C,$$ where $\calN_k^{a}(i)$ is the $k$-neighborhood
defined in (\ref{eq:Nki}) with $\{\bU_i\}$ replaced by $\{\bU_i^a\}$,
and $\calN_k^{b}(i)$ is similarly defined. The cardinality of
$B_{i,j,l}$ is bounded by $k^3$.  Letting $N^*$ be Poisson with rate
$E[N_{d^ad^b}]$, application of the Chen-Stein theorem
\cite[Thm. 1]{arratia1990poisson} yields
\be
\max_A|P(N_{d^ad^b}\in A)-P(N^*\in A)|\leq b_1+b_2+b_3,
\label{eq:chenstein2}
\ee
$$
b_1= \sum_{(i_1,i_2,i_3)\in C} \sum_{(j_1,j_2,j_3)\in B_{i_1,i_2,i_3}} E[\phi_{i_1,i_2,i_3}^{\ab}] E[\phi_{j_1,j_2,j_3}^{\ab}],$$
\be b_2= \sum_{(i_1,i_2,i_3)\in C} \sum_{(j_1,j_2,j_3)\in
  B_{i_1,i_2,i_3}-\{(i_1,i_2,i_3)\}}
E[\phi_{i_1,i_2,i_3}^{\ab}\phi_{j_1,j_2,j_3}^{\ab}],
\label{eq:b2persist}
\ee
$$
b_3 = \sum_{(i_1,i_2,i_3)\in C}
E\left[E\left[\left. \phi_{i_1,i_2,i_3}^{\ab}-p_{i_1,i_2,i_3}^{\ab}\right|
\left\{\phi_{j_1,j_2,j_3}:(j_1,j_2,j_3) \not \in B_{i_1,i_2,i_3}\cup \{i_1,i_2,i_3)\}\right\}\right]\right],$$
with $p_{i_1,i_2,i_3}=E[\phi_{i_1,i_2,i_3}^{\ab}]$.

Next (\ref{eq:phi_ijbnd}) and (\ref{eq:phi_ijbnd2}) are applied to
bound $b_1$ and $b_2$. For $i_1,i_2,i_3\in C$
$$E[\phi_{i_1,i_2,i_3}^{\ab}]= E[\phi_{i_1,i_2}^a]E[\phi_{i_1,i_3}^b] \leq a_{n_a}
a_{n_b}P_{a,0}P_{b,0}M_{1|1}^aM_{1|1}^b.$$
We conclude that
$$ b_1\leq k^3p^3 P_{a,0}^2 P_{b,0}^2 \gamma_0 ,$$
where $\gamma_0=(M_{1|1}^aM_{1|1}^b a_{n_a}a_{n_b})^2$.
Bounding $b_2$ requires more care.  Start from
$$E[\phi_{i_1,i_2,i_3}^{\ab}\phi_{j_1,j_2,j_3}^{\ab}] =
E[\phi_{i_1,i_2}^a\phi_{j_1,j_2}^a]E[\phi_{i_1,i_3}^b\phi_{j_1,j_3}^b],$$
The symmetry relation $\phi_{ij}=\phi_{ji}$ can cause three types of
reductions in the above expression over the range of indices of
summation $i_1, i_2,i_3,j_1,j_2,j_3$ in (\ref{eq:b2persist}). The
first reduction is
$E[\phi_{i_1,i_2}^a\phi_{j_1,j_2}^a]=E[\phi_{i_1,i_2}^a]$, which
occurs when $ i_1 = j_2, i_2= j_1 $, and the second is
$E[\phi_{i_1,i_3}^b\phi_{j_1,j_3}^b]=E[\phi_{i_1,i_3}^b]$, which
occurs when $ i_1 = j_3, i_3= j_1 $. The third reduction occurs when
both of these two reductions occur simultaneously, which is possible
if and only if $i_2=i_3$ and $j_2 = j_3$.  These reductions affect the
order of the summand in $P_{a,0}$ and $P_{b,0}$.  For $i_1 \neq i_2$,
$j_1 \neq j_2$,
$$E[\phi_{i_1,i_2}^a\phi_{j_1,j_2}^a]\leq \left\{\begin{array}{cc}
a_{n_a}P_{a,0} M_{1|1}^a, & i_1 = j_2, i_2= j_1 \\
a_{n_a}^2P_{a,0}^2 \max\{M_{2|1}^a,M_{2|2}^a\} , & o.w.
\end{array}\right.
$$
and similarly for $E[\phi_{i_1,i_3}^b\phi_{j_1,j_3}^b]$.
Hence
\ben b_2 &\leq& p^3 k^3 P_{a,0}^2P_{b,0}^2 \gamma_1 + p^3 k
(P_{a,0}P_{b,0}^2\gamma_2+ P_{a,0}^2P_{b,0}\gamma_3)+ p^2
P_{a,0}P_{b,0}\gamma_4\\
&=& O\left((k/p)^3\right)+ O\left(p^{-1/2}(k/p)\right)+ O\left(p^{-1}\right)
\een
where $\gamma_i$'s are constants depending on
$M_{1|1}^a, M_{2|1}^a,M_{2|2}^a$ and $M_{1|1}^b, M_{2|1}^b,M_{2|2}^b$.  We
conclude that $b_1$ and $b_2$ converge to zero at rates no worse than
$O\left((k/p)^3\right)$ and $O\left(\max\{(k/p)^3,
(k/p)p^{-1/2}, p^{-1}\}\right)$,
respectively.

Finally we deal with the term $b_3$ in (\ref{eq:chenstein2}). Define
$A_k^{\ab}(i_1,i_2,i_3) = (B_{i_1,i_2,i_3}\cup \{(i_1,i_2,i_3))^c$.
Using the definition $\phi_{ijl}^{\ab}=\phi_{ij}^a\phi_{il}^b$ and the
statistical independence of $\phi_{ij}^a$ and $\phi_{il}^b$ the summand of $b_3$ takes the form:
\be &&E[E[\phi_{i_1,i_2,i_3}^{\ab}-p_{i_1,i_2,i_3}|\phi_{A_k^{\ab}(i_1,i_2,i_3)}]]
\nonumber\\&&\hspace{0.5in}=E[E[\phi_{i_1,i_2}^a-p_{i_1,i_2}^a|\bU_{A_k^a(i_1,i_2)}^a]]
E[E[\phi_{i_1,i_3}^b-p_{i_1,i_3}^b|\bU_{A_k^b(i_1,i_3)}^b]]
\nonumber\\&&\hspace{0.6in}+p_{i_1,i_3}^bE[E[\phi_{i_1,i_2}^a-p_{i_1,i_2}^a|\bU_{A_k^a(i_1,i_2)}^a]]+p_{i_1,i_2}^aE[E[\phi_{i_1,i_3}^b-p_{i_1,i_3}^b|\bU_{A_k^b(i_1,i_3)}^b]],
\label{eq:prodpersist}
\ee where $p_{i_1,i_2}^a=E[\phi_{i_1,i_2}^a]$ and $A_k^a(i_1,i_2)$ is
as defined in (\ref{eq:Akij}) for $\bX=\bX^a$ the variables in
treatment $a$. Analogous definitions hold for $p_{i_1,i_3}^b$ and
$A_k^b(i_1,i_3)$. Bounds on the two conditional expectations
the right of (\ref{eq:prodpersist}) were obtained in the proof of
Prop. \ref{prop1}. Using these results in (\ref{eq:prodpersist})
and summing over $(i_1,i_2,i_3)\in C$ yields
$$
|b_3| \leq p^3 P_{a,0}P_{b,0}a_{n_a}a_{n_b} \|\Delta_{p,k}^a\|_2\|\Delta_{p,k}^b\|_2
+p^3(P_{b,0}a_{n_b})^2M_{2|1}^b \|\Delta_{p,k}^a\|_1+p^3(P_{a,0}b_{n_a})^2M_{2|1}^a \|\Delta_{p,k}^b\|_1,
$$
or
$b_3\leq O\left(\max\{\|\Delta_{p,k}^a\|_1,\|\Delta_{p,k}^b\|_1\}\right)$.
Since $k=o(p)$ and the dependency coefficients $\Delta_{p,k}^a, \Delta_{p,k}^b$
converge to zero,  we conclude that $b_1+b_2+b_3$ converge to zero and
therefore $N_{d^ad^b}$ converges in distribution to a Poisson random variable.
This completes the proof of Prop. \ref{prop1persistent}.\qed

\begin{corollaries}
Under the assumptions of Prop. \ref{prop1persistent} the individual
treatment means $p^{-1/2}E[N^a]$ and $p^{-1/2}E[N^b]$ converge to
their respective limits specified in Prop. \ref{prop1}.
\label{corrpersistent}
\end{corollaries}

{\it Proof}: Under the stated conditions in Prop. \ref{prop1persistent} on the
sequences $\rho_p^a$ and $\rho_p^b$,
$p^{1/2}(p-1)P_0(\rho_p^a,n_a)$ and $p^{1/2}(p-1)P_0(\rho_p^b,n_b)$
converge to constants. Furthermore, from the inequality
(\ref{eq:prop1proofpart1}) established in proving Prop. \ref{prop1} (with
$N=N^a,N^b$)
\begin{equation}
\begin{split}
\left| E[N]/\sqrt{p} - \sqrt{p} (p-1) P_0J(\ol{f_{\bfU_{\ast},\bfU_{\ast-\bullet}}}) \right| & \\
\quad \quad \quad \quad \quad \quad \quad  &\leq \left(\gamma_p \left(\sqrt{p}(p-1)P_0\right)^2/\sqrt{p} +
\eta_p (\sqrt{p}(p-1)P_0) \sqrt{2(1-\rho_p)}\right).
\end{split}\label{eq:sartconvbnd}
\end{equation}
and thus $E[N^a]/\sqrt{p}$ and $E[N^b]/\sqrt{p}$ are convergent. This establishes Corollary \ref{corrpersistent}. \qed

We comment on the convergence rates in the three Propositions.  The
dominant distributional convergence rates are identical if the
row-sparse covariance parameter $k$ is fixed but they differ if $k$
increases in $p$.  Assume that the rows of $\mathbb X$ are i.i.d. and
ellipically distributed with a covariance matrix $\mathbf \Sigma$ that
is row-sparse of degree-$k$ with $k=o(p)$. Then for each of the auto-screening,
cross-screening and persistent-screening cases $P(N>0)$ converges to a
Poisson probability of the form $1-\exp(-\Lambda)$ at speed no worse
than $O(p^{-1})$ if $k$ is constant. On the other hand the speed can
be at the slower rates $O\left((k/p)^2\right)$ for auto- and cross-
correlation screening and $O\left((k/p)^3\right)$ for persistent
correlation screening if $k$ increases rapidly with $p$.  On the other
hand the mean number of discoveries may converge to the stated limits
at slower rates. For example, the mean number of auto-correlation
discoveries converges at rate not exceeding
$O(\max\{p^{-1},p^{-2/(n-2)}\})$ while the mean number of persistent
discoveries converges at rate not exceeding
$O(\max\{p^{-1/2},p^{-2/(n-2)}\})$, where $n$ is the minimum of
$n_a,n_b$.

\end{document}